\documentclass{article} 
\usepackage{iclr2017_conference,times}
\usepackage{hyperref}
\usepackage{url}
\usepackage{color}
\usepackage{xcolor}
\usepackage{graphicx}
\usepackage{bm}
\usepackage[export]{adjustbox}
\usepackage{booktabs} 
\usepackage{subcaption}
\usepackage{algpseudocode,algorithm,algorithmicx}
\usepackage{comment}
\usepackage{subcaption}
\usepackage{booktabs, dcolumn}
\newcommand*\Let[2]{\State #1 $\gets$ #2}

\newcommand{\cut}[1]{}

\usepackage[author={Dhruv Batra}]{pdfcomment}

\usepackage{amsmath}

\iclrfinalcopy

\title{LR-GAN: Layered Recursive Generative Adversarial Networks for Image Generation}

\author{Jianwei Yang\thanks{Work was done while visiting Facebook AI Research.} \\
Virginia Tech \\ 
Blacksburg, VA \\
\texttt{jw2yang@vt.edu} \\
\And
Anitha Kannan \\
Facebook AI Research \\
Menlo Park, CA \\
\texttt{akannan@fb.com} \\
\And
Dhruv Batra\footnotemark[1] \ and Devi Parikh\footnotemark[1] \\
Georgia Institute of Technology  \\
Atlanta, GA \\
\texttt{\{dbatra, parikh\}@gatech.edu}
}

\begin{document}

\maketitle

\begin{abstract}


We present LR-GAN: an adversarial image generation model which takes scene structure and context into account. Unlike previous generative adversarial networks (GANs), the proposed GAN learns to generate image background and foregrounds separately and recursively, and stitch the foregrounds on the background in a contextually relevant manner to produce a complete natural image. For each foreground, the model learns to generate its appearance, shape and pose. The whole model is unsupervised, and is trained in an end-to-end manner with gradient descent methods. The experiments demonstrate that LR-GAN can generate more natural images with objects that are more human recognizable than DCGAN. The code is available at \url{https://github.com/jwyang/lr-gan.pytorch}.
\end{abstract}

\section{Introduction}
\vspace*{-7pt}
Generative adversarial networks (GANs)~\citep{GAN} have shown significant promise 
as generative models for natural images. 
A flurry of recent work has proposed improvements over the original GAN work for 
image generation~\citep{DCGAN, LAPGAN, ImprovedGAN, InfoGAN, ManipulateGAN, EBGAN}, multi-stage image generation including part-based models~\citep{GRAN, CompositeGAN}, image generation conditioned on input text or attributes~\citep{CapGAN, Text2ImgGAN, WWGAN}, image generation based on 3D structure~\citep{StyleGAN}, and even video generation \citep{VideoGAN}.

While the holistic `gist' of images generated by these approaches is beginning to look natural, 
there is clearly a long way to go. 
For instance, the foreground objects in these images tend to be 
deformed, blended into the background, and not look realistic or recognizable. 

One fundamental limitation of these methods is that they attempt to generate images without taking into account that images are 2D projections of a 3D visual world, which has a lot of structures in it. This manifests as structure in the 2D images that capture this world. One example of this structure is that images tend to have a background, and foreground objects are placed in this background in contextually relevant ways.

We develop a GAN model that explicitly encodes this structure. Our proposed model generates images in a recursive fashion: it first generates a background, and then conditioned on the background generates a foreground along with a shape (mask) and a pose (affine transformation) that together define how the background and foreground should be composed to obtain a complete image. Conditioned on this composite image, a second foreground and an associated shape and pose are generated, and so on. As a byproduct in the course of recursive image generation, our approach generates some object-shape foreground-background masks in a completely unsupervised way, without access to \emph{any} object masks for training. Note that decomposing a scene into foreground-background layers is a classical ill-posed problem in computer vision. By explicitly factorizing appearance and transformation, LR-GAN encodes natural priors about the images that the same foreground can be `pasted' to the different backgrounds, under different affine transformations. According to the experiments, the absence of these priors result in degenerate foreground-background decompositions, and thus also degenerate final composite images.

\begin{figure}[t]
\begin{minipage}{0.245\linewidth}
\center
\includegraphics[scale=0.251]{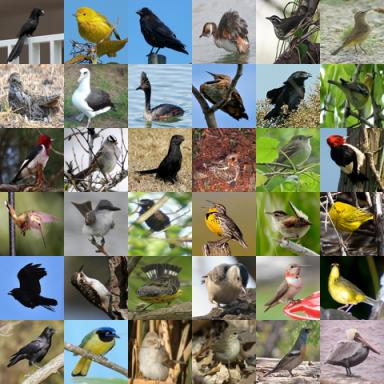}
\vspace{-0.3cm}
\end{minipage}
\begin{minipage}{0.245\linewidth}
\center
\includegraphics[scale=0.251]{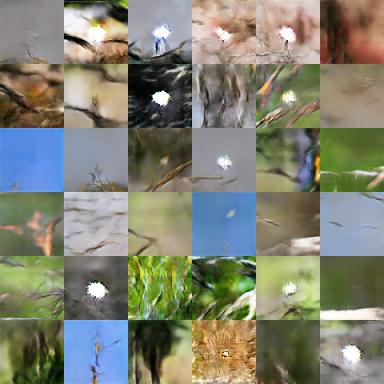}
\vspace{-0.3cm}
\end{minipage}
\begin{minipage}{0.245\linewidth}
\center
\includegraphics[scale=0.251]{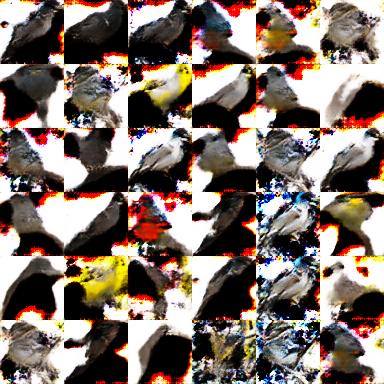}
\vspace{-0.3cm}
\end{minipage}
\begin{minipage}{0.245\linewidth}
\center
\includegraphics[scale=0.251]{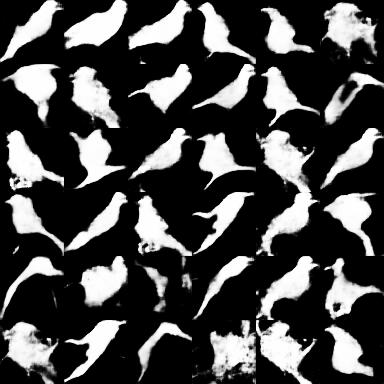}
\vspace{-0.3cm}
\end{minipage}
\begin{subfigure}{0.245\linewidth}
\center
\includegraphics[scale=0.251]{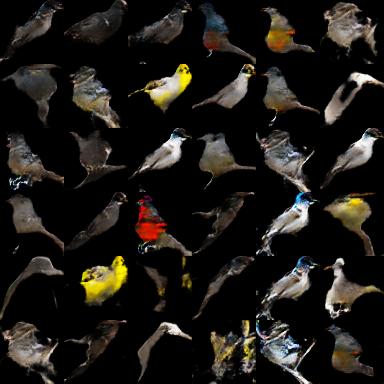}
\end{subfigure}
\begin{subfigure}{0.245\linewidth}
\center
\includegraphics[scale=0.251]{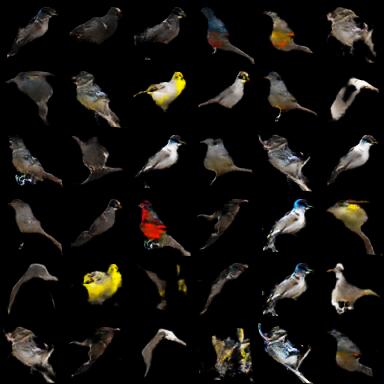}
\end{subfigure}
\begin{subfigure}{0.245\linewidth}
\center
\includegraphics[scale=0.251]{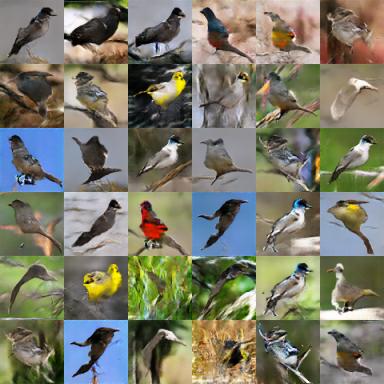}
\end{subfigure}
\begin{subfigure}{0.245\linewidth}
\center
\includegraphics[scale=0.251]{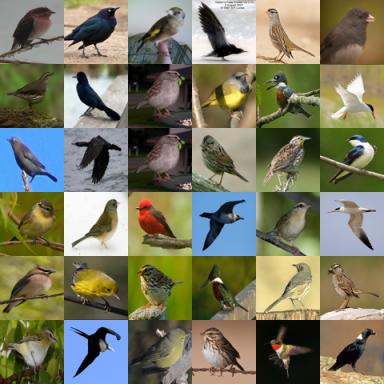}
\end{subfigure}
\caption{Generation results of our model on CUB-200 \citep{CUB200}. It generates images in two timesteps. At the first timestep, it generates background images, while generates foreground images, masks and transformations at the second timestep. Then, they are composed to obtain the final images. From top left to bottom right (row major), the blocks are real images, generated background images, foreground images, foreground masks, carved foreground images, carved and transformed foreground images, final composite images, and their nearest neighbor real images in the training set. Note that the model is trained in a completely unsupervised manner.}
\label{Fig_Intro_CUB200}
\end{figure}

We mainly evaluate our approach on four datasets: MNIST-ONE (one digit) and MNIST-TWO (two digits) synthesized from MNIST \citep{MNIST}, CIFAR-10 \citep{CIFAR10} and CUB-200 \citep{CUB200}. We show qualitatively (via samples) and quantitatively (via evaluation metrics and human studies on Amazon Mechanical Turk) that LR-GAN generates images that globally look natural \emph{and} contain clear background and object structures in them that are realistic and recognizable by humans as semantic entities. An experimental snapshot on CUB-200 is shown in Fig.~\ref{Fig_Intro_CUB200}. We also find that LR-GAN generates foreground objects that are contextually relevant to the backgrounds (e.g., horses on grass, airplanes in skies, ships in water, cars on streets, etc.). For quantitative comparison, besides existing metrics in the literature, we propose two new quantitative metrics to evaluate the quality of generated images. The proposed metrics are derived from the sufficient conditions for the closeness between generated image distribution and real image distribution, and thus supplement existing metrics.

\section{Related Work}
\vspace*{-7pt}
Early work in parametric texture synthesis was based on a set of hand-crafted features \citep{portilla2000parametric}.  Recent improvements in image generation using deep neural networks mainly fall into one of the two stochastic models: variational autoencoders (VAEs) \citep{KSW+16} and generative adversarial networks (GANs) \citep{GAN}. VAEs pair a top-down probabilistic generative network with a bottom up recognition network for amortized probabilistic inference. Two networks are jointly trained to maximize a variational lower bound on the data likelihood. GANs consist of a generator and a discriminator in a minmax game with the generator aiming to fool the discriminator with its samples with the latter aiming to not get fooled.  

Sequential models have been pivotal for improved image generation using variational autoencoders: DRAW \citep{GDG+15} uses attention based recurrence conditioning on the canvas drawn so far. In \cite{EHWTKH16}, a recurrent generative model that draws one object at a time to the canvas was used as the decoder in VAE. These methods are yet to show scalability to natural images. Early compelling results using GANs used sequential coarse-to-fine multiscale generation and class-conditioning \citep{LAPGAN}. Since then, improved training schemes \citep{ImprovedGAN} and better convolutional structure \citep{DCGAN} have improved the generation results using GANs. PixelRNN \citep{pixelRNN} is also recently proposed to sequentially generates a pixel at a time, along the two spatial dimensions.

In this paper, we combine the merits of sequential generation with the flexibility of GANs. Our model for sequential generation imbibes a recursive structure that more naturally mimics image composition by inferring three components: appearance, shape, and pose. One closely related work combining recursive structure with GAN is that of \cite{GRAN} but it does not explicitly model object composition and follows a similar paradigm as by \cite{GDG+15}. \textcolor{black}{Another closely related work is that of \cite{CompositeGAN}. It combines recursive structure and alpha blending. However, our work differs in three main ways: (1) We explicitly use a generator for modeling the foreground poses. That provides significant advantage for natural images, as shown by our ablation studies; (2) Our shape generator is separate from the appearance generator. This factored representation allows more flexibility in the generated scenes; (3) Our recursive framework generates subsequent objects conditioned on the current and previous hidden vectors, \emph{and} previously generated object. This allows for explicit contextual modeling among generated elements in the scene. See Fig.~\ref{Fig_CIFAROutputs_fixbg} for contextually relevant foregrounds generated for the same background, or Fig.~\ref{Fig_MNISTDisentangle} for meaningful placement of two MNIST digits relative to each.}

Models that provide supervision to image generation using conditioning variables have also been proposed: Style/Structure GANs \citep{StyleGAN} learns separate generative models for style and structure that are then composed to obtain final images. In \cite{WWGAN}, GAN based image generation is conditioned on text and the region in the image where the text manifests, specified during training via keypoints or bounding boxes. While not the focus of our work, the model proposed in this paper can be easily extended to take into account these forms of supervision.

\section{Preliminaries}

\subsection{Generative Adversarial Networks }
\label{sec:gan}
Generative Adversarial Networks (GANs) consist of a generator $G$ and a discriminator $D$ that 
are simultaneously trained with competing goals: The generator $G$ is trained to generate samples that can `fool' a discriminator $D$, while the discriminator is trained to classify its inputs as either real (coming from the training dataset) or fake (coming from the samples of $G$). This competition leads to a minmax formulation with a value function:
\begin{equation}
\min_{\theta_G} \max_{\theta_D} \Big( \mathrm{E}_{\bm{x} \sim p_{data}(x) }[\log(D(\bm{x}; \theta_D))] + \mathrm{E}_{\bm{z} \sim p_z(z)}[\log(1 - D(G(\bm{z}; \theta_G); \theta_D))] \Big),
\label{Eq_GAN}
\end{equation}
where $\bm{z}$ is a random vector from a standard multivariate Gaussian or a uniform distribution $p_z(z)$, $G(\bm{z}; \theta_G)$ maps $\bm{z}$ to the data space, $D(\bm{x})$ is the probability that $\bm{x}$ is real estimated by $D$. The advantage of the GANs formulation is that it lacks an explicit loss function and instead uses the discriminator to optimize the generative model. The discriminator, in turn, only cares whether the sample it receives is on the data manifold, and not whether it exactly matches a particular training example (as opposed to losses such as MSE). Hence, the discriminator provides a gradient signal \emph{only} when the generated samples do not lie on the data manifold so that the generator can readjust its parameters accordingly. This form of training enables learning the data manifold of the training set and not just optimizing to reconstruct the dataset, as in autoencoder and its variants.

While the GANs framework is largely agnostic to the choice of $G$ and $D$, it is clear that generative models with the `right' inductive biases will be more effective in learning from the gradient information \citep{LAPGAN,GRAN,GDG+15,WWGAN,YanYSL15}. With this motivation, we propose a generator that models image generation via a recurrent process -- in each time step of the recurrence, an object with its own appearance and shape is generated and warped according to a generated pose to compose an image in layers.

\subsection{Layered structure of image}
\label{sec:layer}
\textcolor{black}{An image taken of our 3D world typically contains a layered structure. One way of representing an image layer is by its appearance and shape. As an example, an image $\bm{x}$ with two layers, foreground $\bm{f}$ and background $\bm{b}$ may be factorized as:
\begin{equation} 
\bm{x} = \bm{f}  \odot  \bm{m} +\bm{b} \odot (1-\bm{m}), \label{eq:as}
\end{equation}
where $\bm{m}$ is the mask depicting the shapes of image layers, and $\odot$ the element wise multiplication operator. Some existing methods assume the access to the shape of the object either during training \citep{isola2013scene} or \emph{both} at train and test time \citep{WWGAN,YanYSL15}. Representing images in layered structure is even straightforward for video with moving objects \citep{TP16, WA94, KJF05}. \cite{VideoGAN} generates videos by separately generating a fixed background and moving foregrounds. A similar way of generating single image can be found in \cite{CompositeGAN}.}

\textcolor{black}{Another way is modeling the layered structure with object appearance and pose as:
\begin{equation} 
\bm{x} = ST(\bm{f}, \bm{a})+ \bm{b}, \label{eq:ap} 
\end{equation}
where $\bm{f}$ and $\bm{b}$ are foreground and background, respectively; $\bm{a}$ is the affine transformation; $ST$ is the spatial transformation operator. Several works fall into this group \citep{RHS+11, HM15, EHWTKH16}. In \cite{HM15}, images are decomposed into layers of objects with specific poses in a variational autoencoder framework, while the number of objects (i.e., layers) is adaptively estimated in \cite{EHWTKH16}.}

\textcolor{black}{To contrast with these works, LR-GAN uses a layered composition, and the foreground layers simultaneously model all three dominant factors of variation: appearance $\bm{f}$, shape $\bm{m}$ and pose $\bm{a}$. We will elaborate it in the following section.}

\section{Layered  Recursive GAN (LR-GAN)} 
\label{sec:model}
\vspace{-5pt}
The basic structure of LR-GAN is similar to GAN: it consists of a discriminator and a generator that are simultaneously trained using the minmax formulation of GAN, as described in \S.\ref{sec:gan}. The key innovation of our work is the layered recursive generator, which is what we describe in this section. 

The generator in LR-GAN is recursive in that the image is constructed recursively 
using a recurrent network. Layered in that each recursive step composes an object layer that is 
`pasted' on the image generated so far. Object layer at timestep $t$ is parameterized by the following three constituents -- `canonical' appearance $\bm{f}_t$, shape (or mask) $\bm{m}_t$, and pose (or affine transformation) $\bm{a}_t$ for warping the object before pasting in the image composition. 

Fig.~\ref{Fig_FlowChart} shows the  architecture of the LR-GAN with the generator architecture 
unrolled for generating background $\bm{x}_0$ ($\doteq \bm{x}_b$) 
and foreground $\bm{x}_1$ and $\bm{x}_2$. At each time step $t$, the generator composes the next image $\bm{x}_t$ via the following recursive computation: 
\begin{equation}
\bm{x}_t = \underbrace{ST(\bm{m}_t,\bm{a}_t)}_{\text{affine transformed mask}} 
\odot \underbrace{ST(\bm{f}_t,\bm{a}_t)}_{\text{affine transformed appearance}} + 
\underbrace{(1- ST(\bm{m}_t,\bm{a}_t)) \odot \bm{x}_{t-1}}_{\text{pasting on image composed so far}}, 
\quad \forall t \in [1,T]
\label{eqn:pasting}
\end{equation}
where $ST(\diamond, \bm{a}_t)$ is a spatial transformation operator that outputs the affine transformed version of $\diamond$ with $\bm{a}_t$ indicating parameters of the affine transformation.

Since our proposed model has an explicit transformation variable $\bm{a}_t$ that is used to warp the object, it can learn a canonical object representation that can be re-used to generate scenes where the object occurs as mere transformations of it, such as different scales or rotations. By factorizing the appearance, shape and pose, the object generator can focus on separately capturing regularities in these three factors that constitute an object. We will demonstrate in our experiments that removing these factorizations from the model leads to its spending capacity in variability that may not solely be about the object in Section \ref{sec:transformation} and \ref{sec:shape}.

\subsection{Details of Generator architecture}
\vspace{-5pt}
\begin{figure}[t]
\center
\includegraphics[scale=0.28]{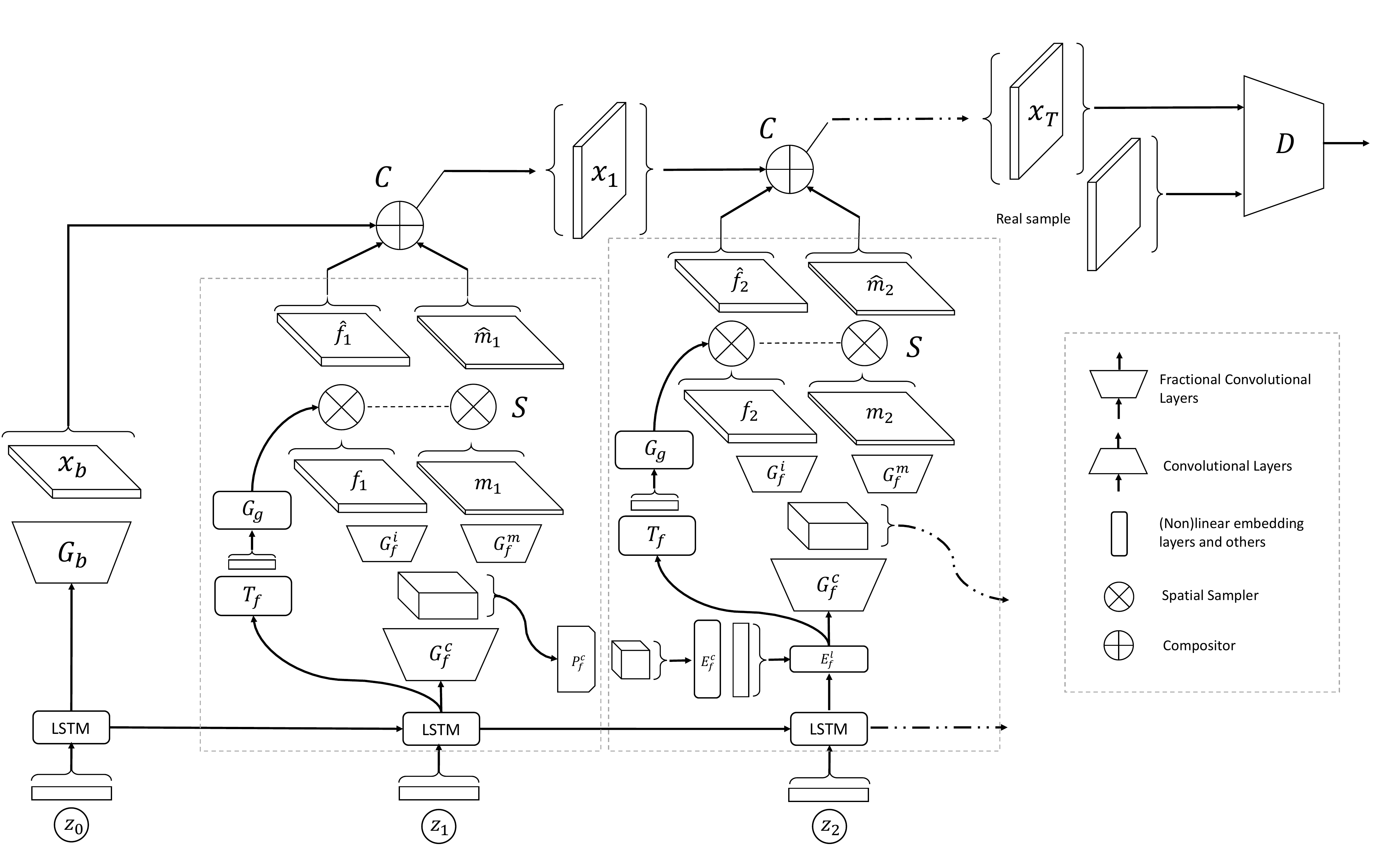}
\caption{LR-GAN architecture unfolded to three timesteps. It mainly consists of one background generator, one foreground generator, temporal connections and one discriminator. The meaning of each component is explained in the legend.} 
\label{Fig_FlowChart}
\end{figure}

Fig.~\ref{Fig_FlowChart} shows our LR-GAN architecture in detail -- we use different shapes to indicate different kinds of layers (convolutional, fractional convolutional, (non)linear, etc), 
as indicated by the legend. Our model consists of two main pieces -- a background generator $G_b$ and a foreground generator $G_f$. $G_b$ and $G_f$ do not share parameters with each other. $G_b$ computation happens only once, while $G_f$ is recurrent over time, i.e., all object generators share the same parameters. In the following, we will introduce each module and connections between them. 

\textbf{Temporal Connections}.  
LR-GAN has two kinds of temporal connections -- informally speaking, one on `top' and one on `bottom'. The `top' connections perform the act of sequentially `pasting' object layers (Eqn.~\ref{eqn:pasting}). The `bottom' connections are constructed by a LSTM on the noise vectors $\bm{z}_0, \bm{z}_1, \bm{z}_2$. Intuitively, this noise-vector-LSTM provides information to the foreground generator about what else has been generated in past. Besides, when generating multiple objects, we use a pooling layer $P_f^c$ and a fully-connected layer $E_f^c$ to extract the information from previous generated object response map. By this way, the model is able to `see' previously generated objects.

\textbf{Background Generator}. The background generator $G_b$ is purposely kept simple. 
It takes the hidden state of noise-vector-LSTM $h_l^0$ as the input and passes it to a number of fractional convolutional layers (also called `deconvolution' layer in some papers) to generate images at its end. The output of background generator $\bm{x}_b$ will be used as the canvas for the following generated foregrounds.

\textbf{Foreground Generator}. The foreground generator $G_f$ is used to generate an object with appearance and shape. Correspondingly, $G_f$ consists of three sub-modules, $G_f^c$, which is a common `trunk' whose outputs are shared by $G_f^i$ and $G_f^m$. $G_f^i$ is used to generate the foreground appearance $\bm{f}_t$, while $G_f^m$ generates the mask $\bm{m}_t$ for the foreground. All three sub-modules consists of one or more fractional convolutional layers combined with batch-normalization and nonlinear layers. The generated foreground appearance and mask have the same spatial size as the background. The top of $G_f^m$ is a sigmoid layer in order to generate one channel mask whose values range in $(0, 1)$.

\textbf{Spatial Transformer}. To spatially transform foreground objects, 
we need to estimate the transformation matrix. As in \cite{STN}, we predict the affine transformation matrix with a linear layer $T_f$ that has six-dimensional outputs. 
Then based on the predicted transformation matrix, we use a grid generator $G_g$ to generate the corresponding sampling coordinates in the input for each location at the output. 
The generated foreground appearance and mask share the same transformation matrix, and thus the same sampling grid. Given the grid, the sampler $S$ will simultaneously sample the $ \bm{f}_t$ and $\bm{m}_t$ to obtain $\hat{\bm{f}}_t$ and $\hat{\bm{m}}_t$, respectively. Different from \cite{STN}, our sampler here normally performs downsampling, since the the foreground typically has smaller size than the background. Pixels in $\hat{\bm{f}}_t$ and $\hat{\bm{m}}_t$ that are from outside the extent of ${\bm{f}}_t$ and ${\bm{m}}_t$ are set to zero. Finally, $\hat{\bm{f}}_t$ and $\hat{\bm{m}}_t$ are sent to the compositor $C$ which combines the canvas $\bm{x}_{t-1}$ and $\hat{\bm{f}}_t$ through layered composition with blending weights given by $\hat{\bm{m}}_t$ (Eqn.~\ref{eqn:pasting}).

Pseudo-code for our approach and detailed model configuration are provided in the Appendix. 

\subsection{New Evaluation Metrics}
\vspace{-5pt}
Several metrics have been proposed to evaluate GANs, such as Gaussian parzen window \citep{GAN}, Generative Adversarial Metric (GAM) \citep{GRAN} and Inception Score \citep{ImprovedGAN}. The common goal is to measure the similarity between the generated data distribution $P_g(\bm{x}) = G(\bm{z}; \theta_z)$ and the real data distribution $P(\bm{x})$.  Most recently, Inception Score has been used in several works \citep{ImprovedGAN, EBGAN}. However, it is an assymetric metric and could be easily fooled by generating centers of data modes.

In addition to these metrics, we present two new metrics based on the following intuition -- a sufficient (but not necessary) condition for closeness of $P_g(\bm{x})$ and $P(\bm{x})$ is closeness of $P_g(\bm{x}|y)$ and $P(\bm{x}|y)$, i.e., distributions of generated data and real data conditioned on all possible variables of interest $y$, e.g., category label. One way to obtain this variable of interest $y$ is via human annotation. Specifically, given the data sampled from $P_g(\bm{x})$ and $P(\bm{x})$, we ask people to label the category of the samples according to some rules. Note that such human annotation is often easier than comparing samples from the two distributions (e.g., because there is no 1:1 correspondence between samples to conduct forced-choice tests). 

After the annotations, we need to verify whether the two distributions are similar in each category. Clearly, directly comparing the distributions $P_g(\bm{x}|y)$ and $P(\bm{x}|y)$ may be as difficult as comparing $P_g(\bm{x})$ and $P(\bm{x})$. Fortunately, we can use Bayes rule and alternatively compare $P_g(y|\bm{x})$ and $P(y|\bm{x})$, which is a much easier task. In this case, we can simply train a discriminative model on the samples from $P_g(\bm{x})$ and $P(\bm{x})$ together with the human annotations about categories of these samples. With a slight abuse of notation, we use $P_g(y|\bm{x})$ and $P(y|\bm{x})$ to denote probability outputs from these two classifiers (trained on generated samples vs trained on real samples). We can then use these two classifiers to compute the following two evaluation metrics: 

\textbf{Adversarial Accuracy:} Computes the classification accuracies achieved by these two classifiers on a validation set, which can be the training set or another set of real images sampled from $P(\bm{x})$. If $P_g(\bm{x})$ is close to $P(\bm{x})$, we expect to see similar accuracies.

\textbf{Adversarial Divergence:} Computes the KL divergence between $P_g(y|\bm{x})$ and $P(y|\bm{x})$. The lower the adversarial divergence, the closer two distributions are. The low bound for this metric is exactly zero, which means $P_g(y|\bm{x}) = P(y|\bm{x})$ for all samples in the validation set. 

As discussed above, we need human efforts to label the real and generated samples. Fortunately, we can further simplify this. Based on the labels given on training data, we split the training data into categories, and train one generator for each category. With all these generators, we generate samples of all categories. This strategy will be used in our experiments on the datasets with labels given.

\section{Experiment}
\vspace{-5pt}
We conduct qualitative and quantitative evaluations on three datasets: 1) MNIST \citep{MNIST}; 2) CIFAR-10 \citep{CIFAR10}; 3) CUB-200 \citep{CUB200}. To add variability to the MNIST images, we randomly scale (factor of 0.8 to 1.2) and rotate ($-\frac{\pi}{4}$ to $\frac{\pi}{4}$) the digits and then stitch them to $48 \times 48$ uniform backgrounds with random grayscale value between [0, 200]. Images are then rescaled back to $32 \times 32$. Each image thus has a different background grayscale value and a different transformed digit as foreground. We rename this sythensized dataset as \textbf{MNIST-ONE} (single digit on a gray background). We also synthesize a dataset \textbf{MNIST-TWO} containing two digits on a grayscale background. We randomly select two images of digits and perform similar transformations as described above, and put one on the left and the other on the right side of a $78 \times 78$ gray background. We resize the whole image to $64 \times 64$.

We develop LR-GAN based on open source code\footnote{https://github.com/soumith/dcgan.torch}. We assume the number of objects is known. Therefore, for MNIST-ONE, MNIST-TWO, CIFAR-10, and CUB-200, our model has two, three, two, and two timesteps, respectively. Since the size of foreground object should be smaller than that of canvas, we set the minimal allowed scale \footnote{Scale corresponds to the size of the target canvas with respect to the object -- the larger the scale, the larger the canvas, and the smaller the relative size of the object in the canvas. 1 means the same size as the canvas.} in affine transforamtion to be 1.2 for all datasets except for MNIST-TWO, which is set to 2 (objects are smaller in MNIST-TWO). In LR-GAN, the background generator and foreground generator have similar architectures. One difference is that the number of channels in the background generator is half of the one in the foreground generator. We compare our results to that of DCGAN \citep{DCGAN}. Note that LR-GAN without LSTM at the first timestep corresponds exactly to the DCGAN. This allows us to run controlled experiments. In both generator and discriminator, all the (fractional) convolutional layers have $4\times4$ filter size with stride 2. As a result, the number of layers in the generator and discriminator automatically adapt to the size of training images. Please see the Appendix (Section \ref{Appendix_ModelConfig}) for details about the configurations. We use three metrics for quantitative evaluation, including Inception Score \citep{ImprovedGAN} and the proposed Adversarial Accuracy, Adversarial Divergence. Note that we report two versions of Inception Score. One is based on the pre-trained Inception net, and the other one is based on the pre-trained classifier on the target datasets.

\subsection{Qualitative Results}
\begin{figure}[t]
 \begin{minipage}{1\textwidth}
    \center
\includegraphics[scale=0.52]{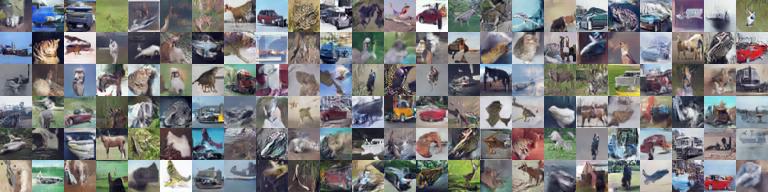}
  \end{minipage}    
  \caption{Generated images on CIFAR-10 based on our model.}
\label{Fig_CIFAR_QualitativeResults}
\end{figure}

\begin{figure}[t]
  \center
  \begin{minipage}{1.0\textwidth}
  \includegraphics[width=0.0620\textwidth]{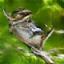}
  \includegraphics[width=0.0620\textwidth]{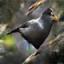}
  \includegraphics[width=0.0620\textwidth]{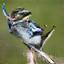}
  \includegraphics[width=0.0620\textwidth]{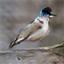}
  \includegraphics[width=0.0620\textwidth]{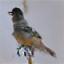}
  \includegraphics[width=0.0620\textwidth]{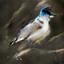}
  \includegraphics[width=0.0620\textwidth]{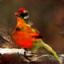}
  \includegraphics[width=0.0620\textwidth]{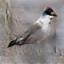}
  \includegraphics[width=0.0620\textwidth]{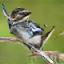}
  \includegraphics[width=0.0620\textwidth]{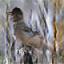}
  \includegraphics[width=0.0620\textwidth]{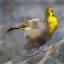}
  \includegraphics[width=0.0620\textwidth]{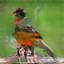}
  \includegraphics[width=0.0620\textwidth]{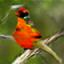}
  \includegraphics[width=0.0620\textwidth]{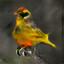}
  \includegraphics[width=0.0620\textwidth]{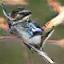}
  \vspace{-0.35cm}
  \end{minipage}     
  
  \begin{minipage}{1.0\textwidth}
  \includegraphics[width=0.0620\textwidth]{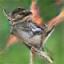}
  \includegraphics[width=0.0620\textwidth]{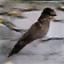}
  \includegraphics[width=0.0620\textwidth]{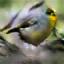}
  \includegraphics[width=0.0620\textwidth]{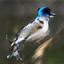}
  \includegraphics[width=0.0620\textwidth]{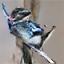}
  \includegraphics[width=0.0620\textwidth]{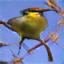}
  \includegraphics[width=0.0620\textwidth]{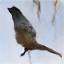}
  \includegraphics[width=0.0620\textwidth]{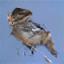}
  \includegraphics[width=0.0620\textwidth]{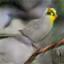}
  \includegraphics[width=0.0620\textwidth]{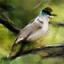}
  \includegraphics[width=0.0620\textwidth]{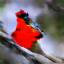}
  \includegraphics[width=0.0620\textwidth]{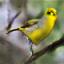}
  \includegraphics[width=0.0620\textwidth]{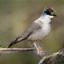}
  \includegraphics[width=0.0620\textwidth]{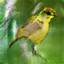}
  \includegraphics[width=0.0620\textwidth]{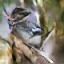}
  \vspace{-0.35cm}  
  \end{minipage}   
   
  \begin{minipage}{1.0\textwidth}
  \includegraphics[width=0.0620\textwidth]{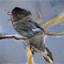}
  \includegraphics[width=0.0620\textwidth]{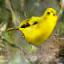}
  \includegraphics[width=0.0620\textwidth]{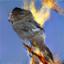}
  \includegraphics[width=0.0620\textwidth]{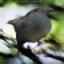}
  \includegraphics[width=0.0620\textwidth]{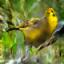}
  \includegraphics[width=0.0620\textwidth]{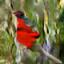}
  \includegraphics[width=0.0620\textwidth]{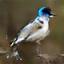}
  \includegraphics[width=0.0620\textwidth]{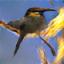}
  \includegraphics[width=0.0620\textwidth]{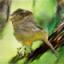}
  \includegraphics[width=0.0620\textwidth]{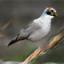}
  \includegraphics[width=0.0620\textwidth]{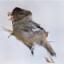}
  \includegraphics[width=0.0620\textwidth]{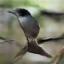}
  \includegraphics[width=0.0620\textwidth]{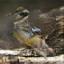}
  \includegraphics[width=0.0620\textwidth]{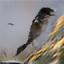}
  \includegraphics[width=0.0620\textwidth]{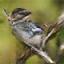}
  \vspace{-0.35cm}  
  \end{minipage}  
  
  \begin{minipage}{1.0\textwidth}
  \includegraphics[width=0.0620\textwidth]{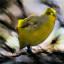}
  \includegraphics[width=0.0620\textwidth]{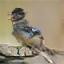}
  \includegraphics[width=0.0620\textwidth]{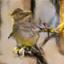}
  \includegraphics[width=0.0620\textwidth]{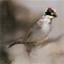}
  \includegraphics[width=0.0620\textwidth]{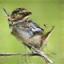}
  \includegraphics[width=0.0620\textwidth]{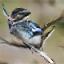}
  \includegraphics[width=0.0620\textwidth]{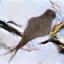}
  \includegraphics[width=0.0620\textwidth]{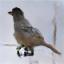}
  \includegraphics[width=0.0620\textwidth]{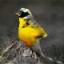}
  \includegraphics[width=0.0620\textwidth]{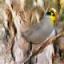}
  \includegraphics[width=0.0620\textwidth]{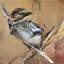}
  \includegraphics[width=0.0620\textwidth]{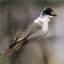}
  \includegraphics[width=0.0620\textwidth]{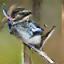}
  \includegraphics[width=0.0620\textwidth]{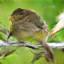}
  \includegraphics[width=0.0620\textwidth]{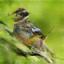}
  \vspace{-0.35cm}  
  \end{minipage}          
\caption{Generated images on CUB-200 based on our model.}
\label{Fig_CUB_QualitativeResults}
\end{figure}

\vspace{-5pt}
In Fig.~\ref{Fig_CIFAR_QualitativeResults} and \ref{Fig_CUB_QualitativeResults}, we show the generated samples for CIFAR-10 and CUB-200, respectively. MNIST results are shown in the next subsection. As we can see from the images, the compositional nature of our model results in the images being free of blending artifacts between backgrounds and foregrounds. For CIFAR-10, we can see the horses and cars with clear shapes. For CUB-200, the bird shapes tend to be even sharper.

\subsection{MNIST-ONE and MNIST-TWO}
\begin{figure}[h]
\center
\includegraphics[scale=0.408]{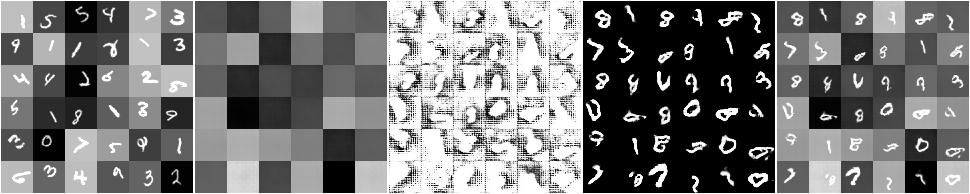}
\caption{Generation results of our model on MNIST-ONE. From left to right, the image blocks are real images, generated background images, generated foreground images, generated masks and final composite images, respectively.}
\label{Fig_MNIST-ONE}
\end{figure}

\vspace{-5pt}
We now report the results on MNIST-ONE and MNIST-TWO. Fig.~\ref{Fig_MNIST-ONE} shows the generation results of our model on MNIST-ONE. As we can see, our model generates the background and the foreground in separate timestep, and can disentagle the foreground digits from background nearly perfectly. Though initial values of the mask randomly distribute in the range of (0, 1), after training, the masks are nearly binary and accurately carve out the digits from the generated foreground. More results on MNIST-ONE (including human studies) can be found in the Appendix (Section~\ref{Appendix_MNIST-ONE}).

Fig.~\ref{Fig_MNISTDisentangle} shows the generation results for MNIST-TWO. Similarly, the model is also able to generate background and the two foreground objects separately. The foreground generator tends to generate a single digit at each timestep. Meanwhile, it captures the context information from the previous time steps. When the first digit is placed to the left side, the second one tends to be placed on the right side, and vice versa. 

\begin{figure}[h]
\begin{minipage}{\linewidth}
\center
\includegraphics[scale=0.255]{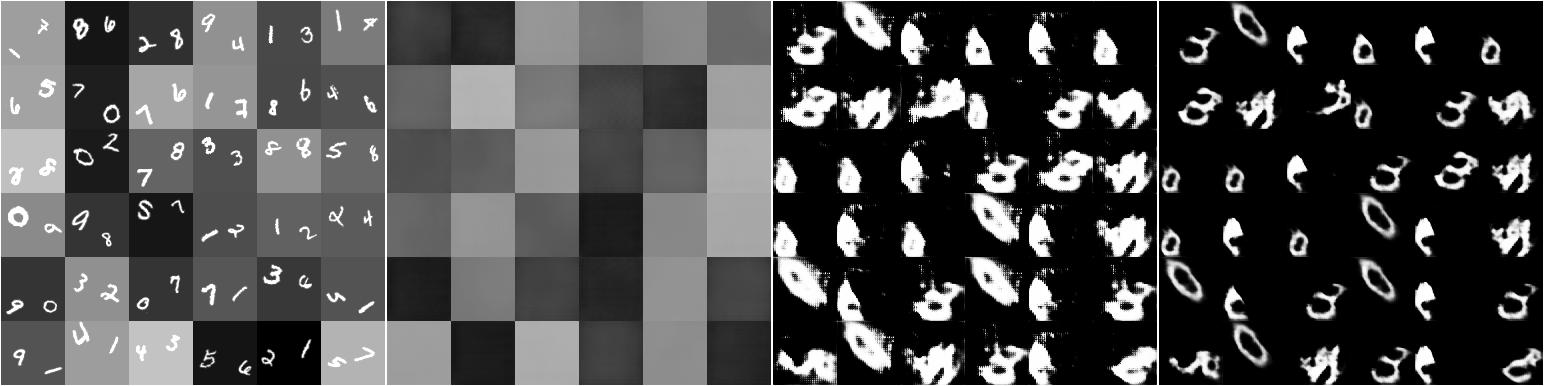}
\end{minipage}
\begin{minipage}{\linewidth}
\center
\includegraphics[scale=0.255]{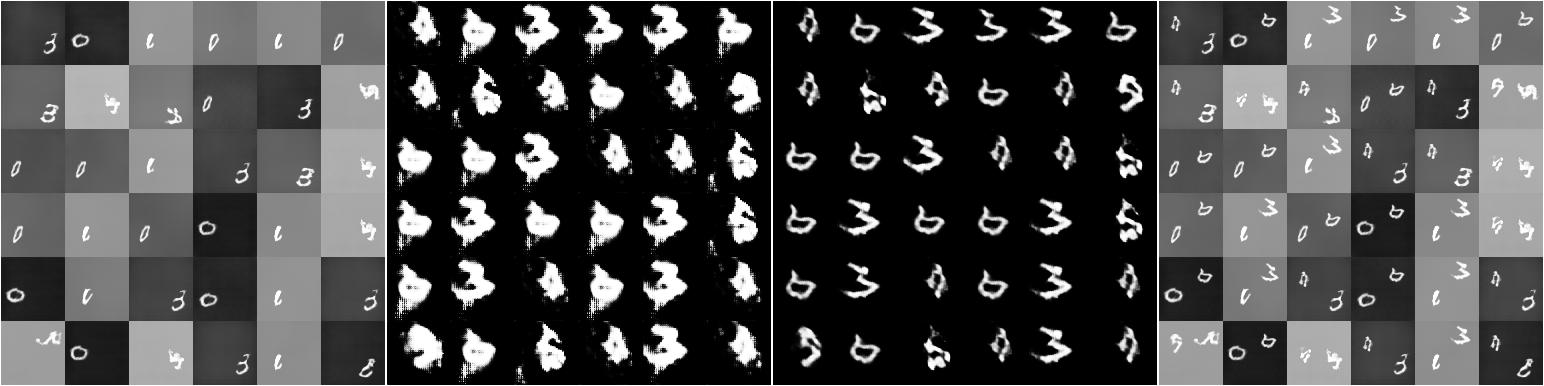}
\end{minipage}
\caption{Generation results of our model on MNIST-TWO. From top left to bottom right (row major), the image blocks are real images, generated background images, foreground images and masks at the second timestep, composite images at the second time step, generated foreground images and masks at the third timestep and the final composite images, respectively.}
\label{Fig_MNISTDisentangle}
\end{figure}

\subsection{CUB-200}
\vspace{-5pt}
\begin{figure}[b]
\begin{minipage}{0.12\linewidth}
\includegraphics[scale=0.335]{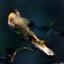}
\includegraphics[scale=0.335]{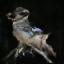}
\vspace{0.05cm}
\end{minipage}
\begin{minipage}{0.12\linewidth}
\includegraphics[scale=0.335]{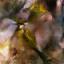}
\includegraphics[scale=0.335]{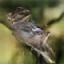}
\vspace{0.05cm}
\end{minipage}
\begin{minipage}{0.12\linewidth}
\includegraphics[scale=0.335]{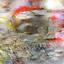}
\includegraphics[scale=0.335]{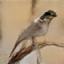}
\vspace{0.05cm}
\end{minipage}
\begin{minipage}{0.12\linewidth}
\includegraphics[scale=0.335]{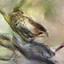}
\includegraphics[scale=0.335]{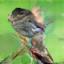}
\vspace{0.05cm}
\end{minipage}
\begin{minipage}{0.12\linewidth}
\includegraphics[scale=0.335]{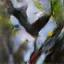}
\includegraphics[scale=0.335]{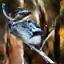}
\vspace{0.05cm}
\end{minipage}
\begin{minipage}{0.12\linewidth}
\includegraphics[scale=0.335]{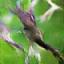}
\includegraphics[scale=0.335]{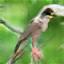}
\vspace{0.05cm}
\end{minipage}
\begin{minipage}{0.12\linewidth}
\includegraphics[scale=0.335]{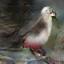}
\includegraphics[scale=0.335]{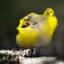}
\vspace{0.05cm}
\end{minipage}
\begin{minipage}{0.12\linewidth}
\includegraphics[scale=0.335]{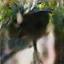}
\includegraphics[scale=0.335]{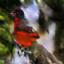}
\vspace{0.05cm}
\end{minipage}

\begin{minipage}{0.12\linewidth}
\includegraphics[scale=0.335]{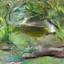}
\includegraphics[scale=0.335]{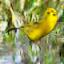}
\vspace{0.05cm}
\end{minipage}
\begin{minipage}{0.12\linewidth}
\includegraphics[scale=0.335]{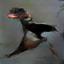}
\includegraphics[scale=0.335]{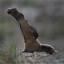}
\vspace{0.05cm}
\end{minipage}
\begin{minipage}{0.12\linewidth}
\includegraphics[scale=0.335]{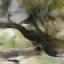}
\includegraphics[scale=0.335]{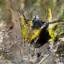}
\vspace{0.05cm}
\end{minipage}
\begin{minipage}{0.12\linewidth}
\includegraphics[scale=0.335]{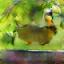}
\includegraphics[scale=0.335]{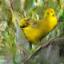}
\vspace{0.05cm}
\end{minipage}
\begin{minipage}{0.12\linewidth}
\includegraphics[scale=0.335]{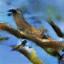}
\includegraphics[scale=0.335]{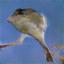}
\vspace{0.05cm}
\end{minipage}
\begin{minipage}{0.12\linewidth}
\includegraphics[scale=0.335]{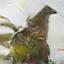}
\includegraphics[scale=0.335]{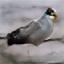}
\vspace{0.05cm}
\end{minipage}
\begin{minipage}{0.12\linewidth}
\includegraphics[scale=0.335]{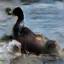}
\includegraphics[scale=0.335]{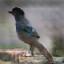}
\vspace{0.05cm}
\end{minipage}
\begin{minipage}{0.12\linewidth}
\includegraphics[scale=0.335]{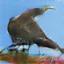}
\includegraphics[scale=0.335]{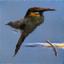}
\vspace{0.05cm}
\end{minipage}

\begin{minipage}{0.12\linewidth}
\includegraphics[scale=0.335]{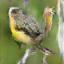}
\includegraphics[scale=0.335]{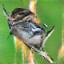}
\vspace{0.05cm}
\end{minipage}
\begin{minipage}{0.12\linewidth}
\includegraphics[scale=0.335]{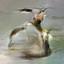}
\includegraphics[scale=0.335]{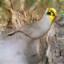}
\vspace{0.05cm}
\end{minipage}
\begin{minipage}{0.12\linewidth}
\includegraphics[scale=0.335]{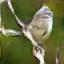}
\includegraphics[scale=0.335]{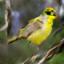}
\vspace{0.05cm}
\end{minipage}
\begin{minipage}{0.12\linewidth}
\includegraphics[scale=0.335]{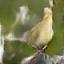}
\includegraphics[scale=0.335]{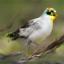}
\vspace{0.05cm}
\end{minipage}
\begin{minipage}{0.12\linewidth}
\includegraphics[scale=0.335]{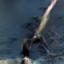}
\includegraphics[scale=0.335]{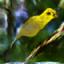}
\vspace{0.05cm}
\end{minipage}
\begin{minipage}{0.12\linewidth}
\includegraphics[scale=0.335]{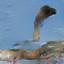}
\includegraphics[scale=0.335]{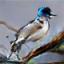}
\vspace{0.05cm}
\end{minipage}
\begin{minipage}{0.12\linewidth}
\includegraphics[scale=0.335]{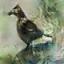}
\includegraphics[scale=0.335]{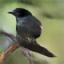}
\vspace{0.05cm}
\end{minipage}
\begin{minipage}{0.12\linewidth}
\includegraphics[scale=0.335]{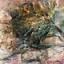}
\includegraphics[scale=0.335]{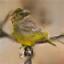}
\vspace{0.05cm}
\end{minipage}

\begin{minipage}{0.12\linewidth}
\includegraphics[scale=0.335]{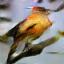}
\includegraphics[scale=0.335]{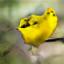}
\vspace{0.05cm}
\end{minipage}
\begin{minipage}{0.12\linewidth}
\includegraphics[scale=0.335]{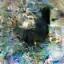}
\includegraphics[scale=0.335]{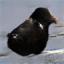}
\vspace{0.05cm}
\end{minipage}
\begin{minipage}{0.12\linewidth}
\includegraphics[scale=0.335]{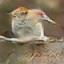}
\includegraphics[scale=0.335]{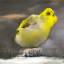}
\vspace{0.05cm}
\end{minipage}
\begin{minipage}{0.12\linewidth}
\includegraphics[scale=0.335]{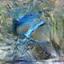}
\includegraphics[scale=0.335]{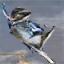}
\vspace{0.05cm}
\end{minipage}
\begin{minipage}{0.12\linewidth}
\includegraphics[scale=0.335]{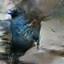}
\includegraphics[scale=0.335]{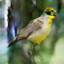}
\vspace{0.05cm}
\end{minipage}
\begin{minipage}{0.12\linewidth}
\includegraphics[scale=0.335]{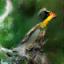}
\includegraphics[scale=0.335]{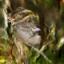}
\vspace{0.05cm}
\end{minipage}
\begin{minipage}{0.12\linewidth}
\includegraphics[scale=0.335]{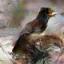}
\includegraphics[scale=0.335]{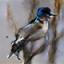}
\vspace{0.05cm}
\end{minipage}
\begin{minipage}{0.12\linewidth}
\includegraphics[scale=0.335]{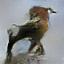}
\includegraphics[scale=0.335]{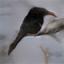}
\vspace{0.05cm}
\end{minipage}

\begin{minipage}{0.12\linewidth}
\includegraphics[scale=0.335]{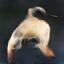}
\includegraphics[scale=0.335]{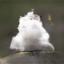}
\vspace{0.05cm}
\end{minipage}
\begin{minipage}{0.12\linewidth}
\includegraphics[scale=0.335]{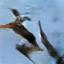}
\includegraphics[scale=0.335]{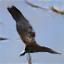}
\vspace{0.05cm}
\end{minipage}
\begin{minipage}{0.12\linewidth}
\includegraphics[scale=0.335]{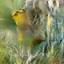}
\includegraphics[scale=0.335]{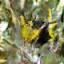}
\vspace{0.05cm}
\end{minipage}
\begin{minipage}{0.12\linewidth}
\includegraphics[scale=0.335]{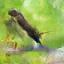}
\includegraphics[scale=0.335]{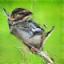}
\vspace{0.05cm}
\end{minipage}
\begin{minipage}{0.12\linewidth}
\includegraphics[scale=0.335]{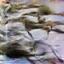}
\includegraphics[scale=0.335]{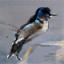}
\vspace{0.05cm}
\end{minipage}
\begin{minipage}{0.12\linewidth}
\includegraphics[scale=0.335]{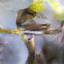}
\includegraphics[scale=0.335]{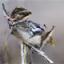}
\vspace{0.05cm}
\end{minipage}
\begin{minipage}{0.12\linewidth}
\includegraphics[scale=0.335]{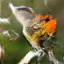}
\includegraphics[scale=0.335]{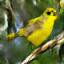}
\vspace{0.05cm}
\end{minipage}
\begin{minipage}{0.12\linewidth}
\includegraphics[scale=0.335]{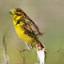}
\includegraphics[scale=0.335]{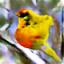}
\vspace{0.05cm}
\end{minipage}
\caption{Matched pairs of generated images based on DCGAN and LR-GAN, respectivedly. The odd columns are generated by DCGAN, and the even columns are generated by LR-GAN. These are paired according to the perfect matching based on Hungarian algorithm.}
\label{Fig_CUBComparison}
\end{figure}
We study the effectiveness of our model trained on the CUB-200 bird dataset. In Fig.~\ref{Fig_Intro_CUB200}, we have shown a random set of generated images, along with the intermediate generation results of the model. While being \emph{completely unsupervised}, the model, for a large fraction of the samples, is able to successfully disentangle the foreground and the background. This is evident from the generated bird-like masks.
%

We do a comparative study based on Amazon Mechanical Turk (AMT) between DCGAN and LR-GAN to quantify relative visual quality of the generated images. We first generated 1000 samples from both the models. Then, we performed perfect matching between the two image sets using the Hungarian algorithm on $L2$ norm distance in the pixel space. This resulted in 1000 image pairs. Some examplar pairs are shown in Fig.~\ref{Fig_CUBComparison}. For each image pair, 9 judges are asked to choose the one that is more realistic. Based on majority voting, we find that our generated images are selected 68.4\% times, compared with 31.6\% times for DCGAN. This demonstrates that our model has generated more realistic images than DCGAN. We can attribute this difference to our model's ability to generate foreground separately from the background, enabling stronger edge cues. 
 

\subsection{CIFAR-10}
\label{exp_cifar10}
\vspace{-5pt}
We now qualitatively and quantitatively evaluate our model on CIFAR-10, which contains multiple object categories and also various backgrounds.
\begin{table}[b]
\center
\caption{Quantitative comparison between DCGAN and LR-GAN on CIFAR-10.}
\small
  \begin{tabular}{l c c c}
    \toprule
    Training Data           & Real Images  & DCGAN  &  Ours \\
    \midrule	    
    Inception Score\textsuperscript{\textdagger}             & 11.18$\pm$0.18      & 6.64$\pm$0.14       & 7.17$\pm$0.07               \\
    Inception Score\textsuperscript{\textdagger\textdagger}           & 7.23$\pm$0.09      & 5.69$\pm$0.07       & 6.11$\pm$0.06       \\
    Adversarial Accuracy                  & 83.33$\pm$0.08      &  37.81$\pm$0.02 & 44.22 $\pm$0.08       \\
    Adversarial Divergence  & 0                 &  7.58$\pm$0.04    & 5.57$\pm$0.06           \\
   \bottomrule
      \multicolumn{4}{l}{\textsuperscript{\textdagger}\footnotesize{Evaluate using the pre-trained Inception net as \cite{ImprovedGAN}}} \\
   \multicolumn{4}{l}{\textsuperscript{\textdagger\textdagger}\footnotesize{Evaluate using the supervisedly trained classifier based on the discriminator in LR-GAN.}}
  \end{tabular}
  \label{Table_CIFAR}
\end{table}
\begin{figure}[t]
\begin{minipage}{0.245\linewidth}
\center
\includegraphics[scale=0.38]{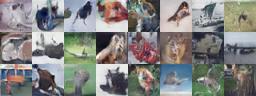}
\end{minipage}
\begin{minipage}{0.245\linewidth}
\center
\includegraphics[scale=0.38]{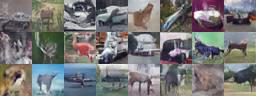}
\end{minipage}
\begin{minipage}{0.245\linewidth}
\center
\includegraphics[scale=0.38]{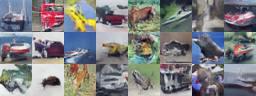}
\end{minipage}
\begin{minipage}{0.245\linewidth}
\center
\includegraphics[scale=0.38]{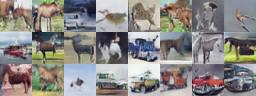}
\end{minipage}

\begin{minipage}{0.245\linewidth}
\center
\includegraphics[scale=0.38]{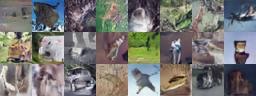}
\end{minipage}
\begin{minipage}{0.245\linewidth}
\center
\includegraphics[scale=0.38]{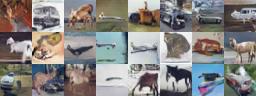}
\end{minipage}
\begin{minipage}{0.245\linewidth}
\center
\includegraphics[scale=0.38]{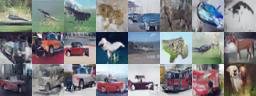}
\end{minipage}
\begin{minipage}{0.245\linewidth}
\center
\includegraphics[scale=0.38]{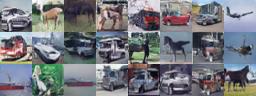}
\end{minipage}
\caption{Qualitative comparison on CIFAR-10. Top three rows are images generated by DCGAN; Bottom three rows are by LR-GAN. From left to right, the blocks display generated images with increasing quality level as determined by human studies.}
\label{Fig_CIFARComparison}
\end{figure}
\begin{figure}[t]
\begin{minipage}{0.138\linewidth}
\center
\includegraphics[scale=0.33]{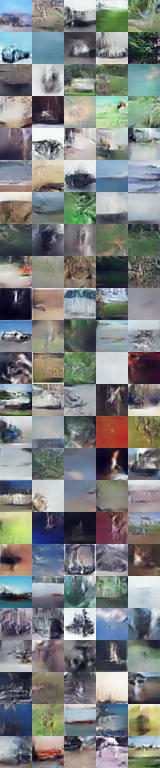}
\end{minipage}
\begin{minipage}{0.138\linewidth}
\center
\includegraphics[scale=0.33]{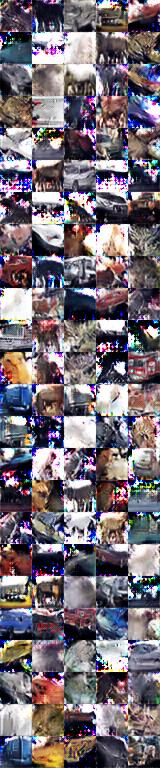}
\end{minipage}
\begin{minipage}{0.138\linewidth}
\center
\includegraphics[scale=0.33]{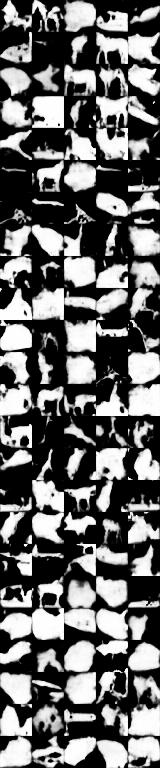}
\end{minipage}
\begin{minipage}{0.138\linewidth}
\center
\includegraphics[scale=0.33]{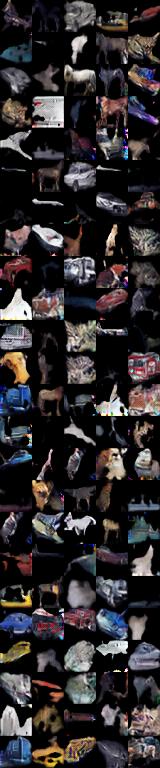}
\end{minipage}
\begin{minipage}{0.138\linewidth}
\center
\includegraphics[scale=0.33]{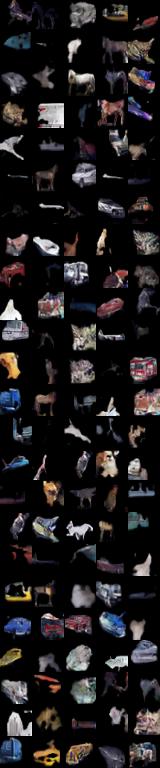}
\end{minipage}
\begin{minipage}{0.138\linewidth}
\center
\includegraphics[scale=0.33]{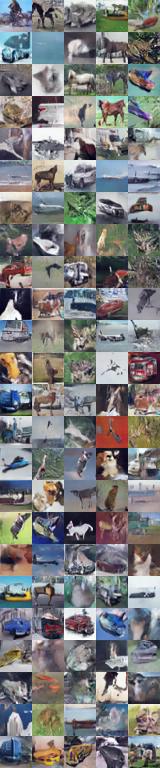}
\end{minipage}
\begin{minipage}{0.138\linewidth}
\center
\includegraphics[scale=0.33]{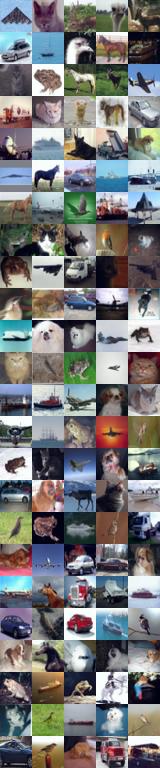}
\end{minipage}
\caption{Generation results of our model on CIFAR-10. From left to right, the blocks are: generated background images, foreground images, foreground masks, foreground images carved out by masks, carved foregrounds after spatial transformation, final composite images and nearest neighbor training images to the generated images.}
\label{Fig_CIFAROutputs_More}
\end{figure} 

\begin{figure}[h]
\begin{minipage}{1\linewidth}
\center
\includegraphics[scale=0.34]{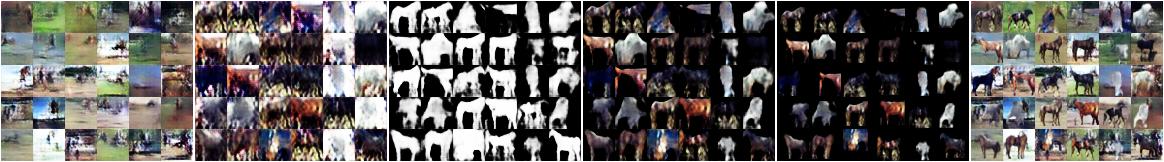}
\end{minipage}
\begin{minipage}{1\linewidth}
\center
\includegraphics[scale=0.34]{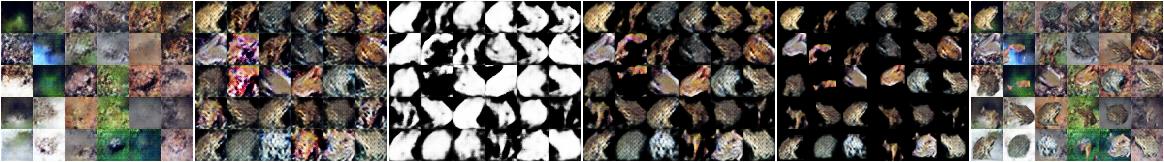}
\end{minipage}
\begin{minipage}{1\linewidth}
\center
\includegraphics[scale=0.34]{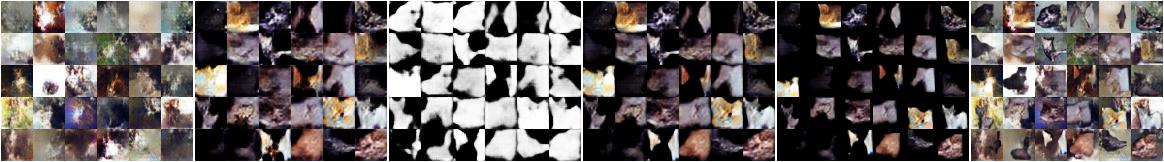}
\end{minipage}
\caption{Category specific generation results of our model on CIFAR-10 categories of horse, frog, and cat (top to bottom). The blocks from left to right are: generated background images, foreground images, foreground masks, foreground images carved out by masks, carved foregrounds after spatial transformation and final composite images.}
\label{Fig_CIFAROutputs_cat_fulll}
\end{figure}

{\bf Comparison of image generation quality:}   We conduct AMT studies to compare the fidelity of image generation.  Towards this goal, we generate 1000 images from DCGAN and LR-GAN, respectively.  We ask 5 judges to label each image to either belong to one of the 10 categories or as  `non recognizable' or `recognizable but not belonging to the listed categories'.  We then assign each image a quality level between [0,5]  that captures the number of judges that agree with the majority choice.  Fig.~\ref{Fig_CIFARComparison} shows the images generated by both  approaches, ordered by increasing quality level. We merge images at quality level 0 (all judges said non-recognizable) and 1 together, and similarly images at level 4 and 5.  Visually, the generated samples by our model have clearer boundaries and object structures.  We also computed the fraction of non-recognizable images: Our model had a 10\% absolute drop in non-recognizability rate (67.3\% for ours vs. 77.7\% for DCGAN). For reference, 11.4\% of real CIFAR images were categorized as non-recognizable. Fig.~\ref{Fig_CIFAROutputs_More} shows more generated (intermediate) results of our model.

{\bf Quantitative evaluation on generators:}  We evaluate the generators based on three metrics: 1) Inception Score; 2) Adversarial Accuracy; 3) Adversarial Divergence. \textcolor{black}{To obtain a classifier model for evaluation, we remove the top layer in the discriminator used in our model, and then append two fully connected layers on the top of it. We train this classifier using the training samples of CIFAR-10 based on the annotations. Following \cite{ImprovedGAN}, we generated 50,000 images based on DCGAN and LR-GAN, repsectively. We compute two types of Inception Scores. The standard Inception Score is based on the Inception net as in \cite{ImprovedGAN}, and the contextual Inception Score is based on our trained classifier model. To distinguish, we denote the standard one as `Inception Score\textsuperscript{\textdagger}', and the contextual one as `Inception Score\textsuperscript{\textdagger\textdagger}'.} To obtain the Adversarial Accuracy and Adversarial Divergence scores, we train one generator on each of 10 categories for DCGAN and LR-GAN, respectively. Then, we use these generators to generate samples of different categories. Given these generated samples, we train the classifiers for DCGAN and LR-GAN separately. Along with the classifier trained on the real samples, we compute the Adversarial Accuracy and Adversarial Divergence on the real training samples. In Table~\ref{Table_CIFAR}, we report the Inception Scores, Adversarial Accuracy and Adversarial Divergence for comparison.  We can see that our model outperforms DCGAN across the board. To point out, we obtan different Inception Scores based on different classifier models, which indicates that the Inception Score varies with different models.

{\bf Quantitative evaluation on discriminators:}  We evaluate the discriminator as an extractor for deep representations. Specifically, we use the output of the last convolutional layer in the discriminator as features. We perform a 1-NN classification on the test set given the full training set. Cosine similarity is used as the metric. 
On the test set, our model achieves 62.09\%$\pm$0.01\% compared to DCGAN's 56.05\%$\pm$0.02\%. 

{\bf Contextual generation:} We also show the efficacy of our approach to generate diverse foregrounds conditioned on fixed background. The results in Fig.~\ref{Fig_CIFAROutputs_fixbg} in Appendix showcase that the foreground generator generates objects that are compatible with the background. This indicates that the model has captured contextual dependencies between the image layers. 

{\bf Category specific models:} The objects in CIFAR-10 exhibit huge variability in shapes. That can partly explain why some of the generated shapes are not as compelling in Fig.~\ref{Fig_CIFAROutputs_More}. To test this hypothesis, we reuse the generators trained for each of 10 categories used in our metrics to obtain the generation results. Fig.~\ref{Fig_CIFAROutputs_cat_fulll} shows results for categories `horse', `frog' and `cat'. We can see that the model is now able to generate object-specific appearances and shapes, similar in vein to our results on the CUB-200 dataset.  

\subsection{Importance of Transformations}
\label{sec:transformation}
\begin{figure}[h]
\begin{minipage}{0.195\linewidth}
\center
\includegraphics[scale=0.4024, cfbox=black 0.1pt 0.1pt]{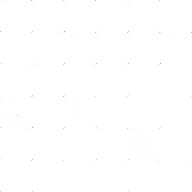}
\end{minipage}
\begin{minipage}{0.195\linewidth}
\center
\includegraphics[scale=0.4024]{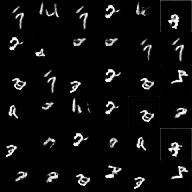}
\end{minipage}
\begin{minipage}{0.195\linewidth}
\center
\includegraphics[scale=0.4024]{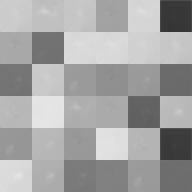}
\end{minipage}
\begin{minipage}{0.195\linewidth}
\center
\includegraphics[scale=0.4024]{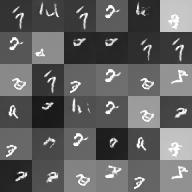}
\end{minipage}
\begin{minipage}{0.195\linewidth}
\center
\includegraphics[scale=0.4024]{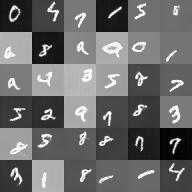}
\end{minipage}

\begin{minipage}{0.195\linewidth}
\center
\includegraphics[scale=0.2012]{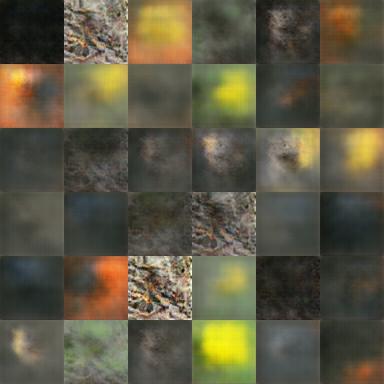}
\end{minipage}
\begin{minipage}{0.195\linewidth}
\center
\includegraphics[scale=0.2012]{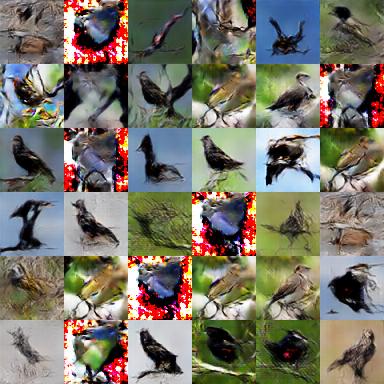}
\end{minipage}
\begin{minipage}{0.195\linewidth}
\center
\includegraphics[scale=0.2012]{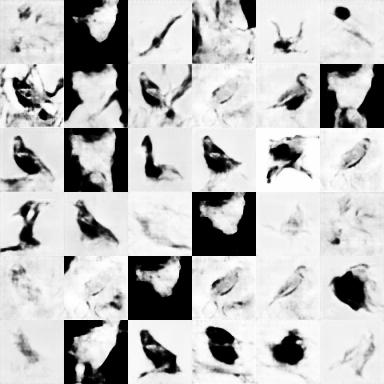}
\end{minipage}
\begin{minipage}{0.195\linewidth}
\center
\includegraphics[scale=0.2012]{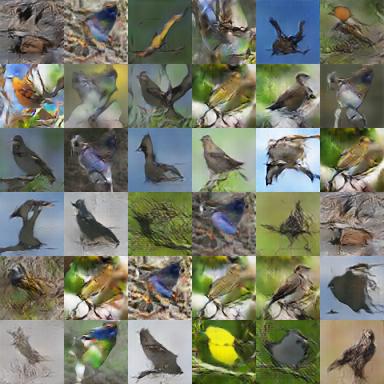}
\end{minipage}
\begin{minipage}{0.195\linewidth}
\center
\includegraphics[scale=0.2012]{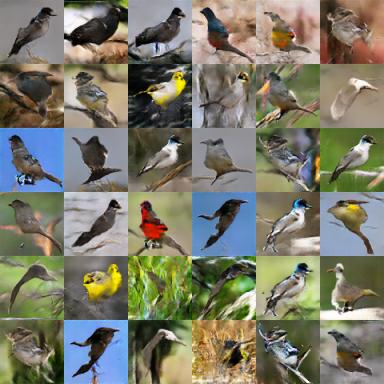}
\end{minipage}

\begin{minipage}{0.195\linewidth}
\center
\includegraphics[scale=0.4]{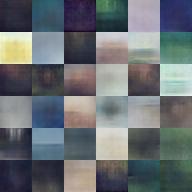}
\end{minipage}
\begin{minipage}{0.195\linewidth}
\center
\includegraphics[scale=0.4]{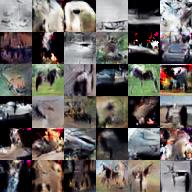}
\end{minipage}
\begin{minipage}{0.195\linewidth}
\center
\includegraphics[scale=0.4]{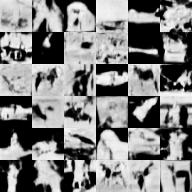}
\end{minipage}
\begin{minipage}{0.195\linewidth}
\center
\includegraphics[scale=0.4]{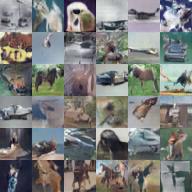}
\end{minipage}
\begin{minipage}{0.195\linewidth}
\center
\includegraphics[scale=0.4]{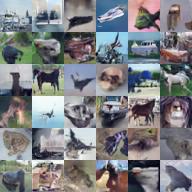}
\end{minipage}
\caption{Generation results from an ablated LR-GAN model without affine transformations. From top to bottom, the block rows correspond to different datasets: MNIST-ONE, CUB-200, CIFAR-10. From left to right, the blocks show generated background images, foreground images, foreground masks, and final composite images. For comparison, the rightmost column block shows final generated images from a non-ablated model with affine transformations.}
\label{Fig_Appendix_noT}
\end{figure}

\textcolor{black}{Fig.~\ref{Fig_Appendix_noT} shows results from an ablated model without affine transformations in the foreground layers, and compares the results with the full model that does include these transformations. We note that one significant problem emerges that the decompositions are degenerate, in the sense that the model is unable to break the symmetry between foreground and background layers, often generating object appearances in the model's background layer and vice versa. For CUB-200, the final generated images have some blendings between foregrounds and backgrounds. This is particularly the case for those images without bird-shape masks. For CIFAR-10, a number of generated masks are inverted. In this case, the background images are carved out as the foreground objects. The foreground generator takes almost all the duty to generate the final images, which make it harder to generate images as clear as the model with transformation. From these comparisons, we qualitatively demonstrate the importance of modeling transformations in the foreground generation process. Another merit of using transformation is that the intermediate outputs of the model are more interpretable and faciliate to the downstreaming tasks, such as scene paring, which is demonstrated in Section~\ref{subsec_CondGeneration}.}

\subsection{Importance of Shapes}
\label{sec:shape}
\begin{figure}[t]
\begin{minipage}{0.195\linewidth}
\center
\includegraphics[scale=0.4]{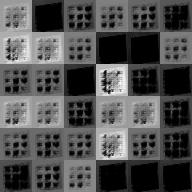}
\end{minipage}
\begin{minipage}{0.195\linewidth}
\center
\includegraphics[scale=0.4]{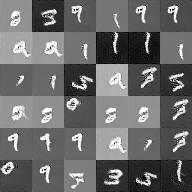}
\end{minipage}
\begin{minipage}{0.195\linewidth}
\center
\includegraphics[scale=0.4]{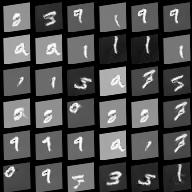}
\end{minipage}
\begin{minipage}{0.195\linewidth}
\center
\includegraphics[scale=0.4]{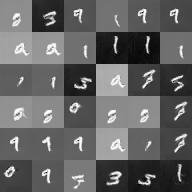}
\end{minipage}
\begin{minipage}{0.195\linewidth}
\center
\includegraphics[scale=0.4]{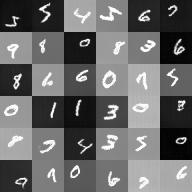}
\end{minipage}

\begin{minipage}{0.195\linewidth}
\center
\includegraphics[scale=0.2012]{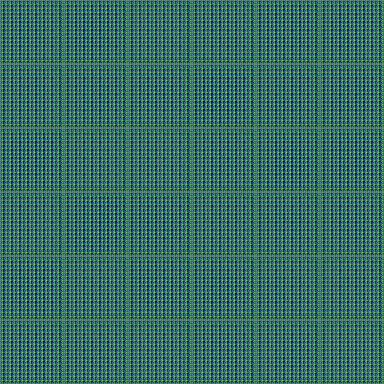}
\end{minipage}
\begin{minipage}{0.195\linewidth}
\center
\includegraphics[scale=0.2012]{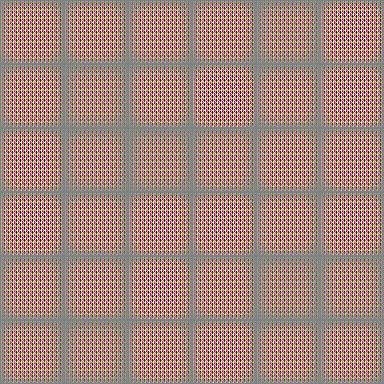}
\end{minipage}
\begin{minipage}{0.195\linewidth}
\center
\includegraphics[scale=0.2012]{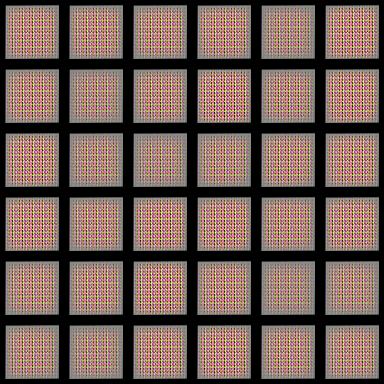}
\end{minipage}
\begin{minipage}{0.195\linewidth}
\center
\includegraphics[scale=0.2012]{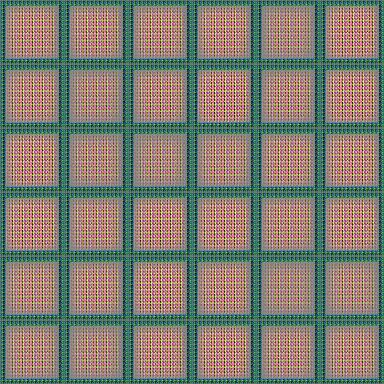}
\end{minipage}
\begin{minipage}{0.195\linewidth}
\center
\includegraphics[scale=0.2012]{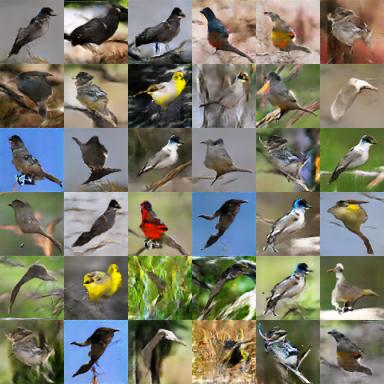}
\end{minipage}

\begin{minipage}{0.195\linewidth}
\center
\includegraphics[scale=0.4]{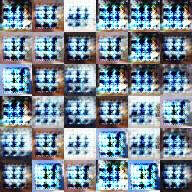}
\end{minipage}
\begin{minipage}{0.195\linewidth}
\center
\includegraphics[scale=0.4]{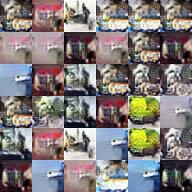}
\end{minipage}
\begin{minipage}{0.195\linewidth}
\center
\includegraphics[scale=0.4]{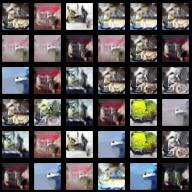}
\end{minipage}
\begin{minipage}{0.195\linewidth}
\center
\includegraphics[scale=0.4]{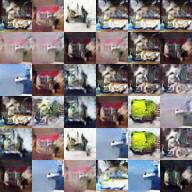}
\end{minipage}
\begin{minipage}{0.195\linewidth}
\center
\includegraphics[scale=0.4]{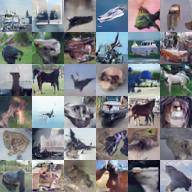}
\end{minipage}
\caption{\textcolor{black}{Generation results from an ablated LR-GAN model without mask generator. The block rows correspond to different datasets (from top to bottom: MNIST-ONE, CUB-200, CIFAR-10). From left to right, the blocks show generated background images, foreground images, transformed foreground images, and final composite images. For comparison, the rightmost column block shows final generated images from a non-ablated model with mask generator.}}
\label{Fig_Appendix_noM}
\end{figure}

\textcolor{black}{We perform another ablation study by removing the mask generator to understand the importance of modeling object shapes. In this case, the generated foreground is simply pasted on top of the generated background after being transformed. There is no alpha blending between the foregrounds and backgrounds. The generation results for three datasets, MNIST-ONE, CUB-200, CIFAR-10 are shown in Fig.~\ref{Fig_Appendix_noM}. As we can see, though the model works well for the generation of MNIST-ONE, it fails to generate reasonable images across the other datasets. Particularly, the training does not even converge for CUB-200. Based on these results, we qualitatively demonstrate that mask generator in our model is fairly important to obtain plausible results, especially for realistic images.}


\bibliography{iclr2017_conference}
\bibliographystyle{iclr2017_conference}

\section{Appendix}
\subsection{Algorithm}
\label{Appendix_Alg}
Algo.~\ref{algorithm} illustrates the generative process in our model. 
$g(\star)$ evaluates the function $g$ at $\star$.  $\circ$ is a composition operator that composes its operands so that $f \circ g(\star) = f(g(\star))$. 

\begin{algorithm}
\caption{Stochastic Layered Recursive Image Generation}
\begin{algorithmic}[1]
\State $\bm{z}_0 \sim \mathcal{N}(0,I)$
\State  $\bm{x}_0 = G_b(\bm{z}_0)$ \Comment{background generator} 
\Let{$\hm{h}_l^0$}{$\bm{0}$}
\Let{$\bm{c}_l^0$}{$\bm{0}$}
\For  {$t \in [1 \cdots T]$ }
\State $\bm{z}_t \sim \mathcal{N}(0,I)$
\Let{$\bm{h}_l^t$, $\bm{c}_l^t$}{LSTM([$\bm{z}_t$, $\bm{h}_l^{t-1}$, $\bm{c}_l^{t-1}$])} \Comment{pass through LSTM} 
\If{t = 1 }
   \Let{$\bm{y}_t$}{$\bm{h}_l^t$}
  \Else
    \Let{$\bm{y}_t$}{$E_f^l$([$\bm{h}_l^t$  $\bm{h}^{{t-1}}_{f}$])} \Comment{pass through non-linear embedding layers $E_f^l$}
 \EndIf
   \Let{$\bm{s}_t$}{ $G_f^{c} (\bm{y}_t)$} \Comment{predict shared cube for $G_f^i$ and $G_f^m$}
           \Let{$a_t$}{$T_f(\bm{y}_t)$} \Comment{ object transformation}
     \Let{$\bm{f}_t$}{ $G_f^{i}  (\bm{s}_t)$} \Comment{ generate object appearance }
        \Let{$\bm{m}_t$}{ $G_f^{m}  (\bm{s}_t)$} \Comment{ generate object shape}
        \Let{ $\bm{h}^{t}_{f}$}{ $E_f^{c}\circ P_f^c (\bm{s}_t)$} \Comment{predict shared represenation embedding }
        \Let{$\bm{x}_t$}{$ST(\bm{m}_t,\bm{a}_t) \odot ST(\bm{f}_t,\bm{a}_t) + (1- ST(\bm{m}_t,\bm{a}_t)) \odot \bm{x}_{t-1}$}
\EndFor
\end{algorithmic}
\label{algorithm}
\end{algorithm}

\subsection{Model Configurations}
\label{Appendix_ModelConfig}
Table~\ref{Tabel_ModelConfig} lists the information and model configuration for different datasets. The dimensions of random vectors and hidden vectors are all set to 100. We also compare the number of parameters in DCGAN and LR-GAN. The numbers before `/' are our model, after `/' are DCGAN. Based on the same notation used in \citep{EBGAN}, the architectures for the different datasets are:
\begin{itemize}
\item MNIST-ONE: $\bm{G}_b$: (256)4c-(128)4c2s-(64)4c2s-(3)4c2s; $\bm{G}_f^c$: (512)4c-(256)4c2s-(128)4c2s; $\bm{G}_f^i$:  (3)4c2s; $\bm{G}_f^m$: (1)4c2s; $D$: (64)4c2s-(128)4c2s-(256)4c2s-(256)4p4s-1

\item MNIST-TWO: $\bm{G}_b$: (256)4c-(128)4c2s-(64)4c2s-(32)4c2s-(3)4c2s; $\bm{G}_f^c$: (512)4c-(256)4c2s-(128)4c2s-(64)4c2s; $\bm{G}_f^i$:  (3)4c2s; $\bm{G}_f^m$: (1)4c2s;$D$: (64)4c2s-(128)4c2s-(256)4c2s-(512)4c2s-(512)4p4s-1

\item CUB-200: $\bm{G}_b$: (512)4c-(256)4c2s-(128)4c2s-(64)4c2s-(3)4c2s; $\bm{G}_f^c$: (1024)4c-(512)4c2s-(256)4c2s-(128)4c2s; $\bm{G}_f^i$:  (3)4c2s; $\bm{G}_f^m$: (1)4c2s;$D$: (128)4c2s-(256)4c2s-(512)4c2s-(1024)4c2s-(1024)4p4s-1

\item CIFAR-10: $\bm{G}_b$: (256)4c-(128)4c2s-(64)4c2s-(3)4c2s; $\bm{G}_f^c$: (512)4c-(256)4c2s-(128)4c2s; $\bm{G}_f^i$:  (3)4c2s; $\bm{G}_f^m$: (1)4c2s $D$: (64)4c2s-(128)4c2s-(256)4c2s-(256)4p4s-1
\end{itemize} 

\begin{table}[t]\footnotesize
\setlength{\tabcolsep}{5.5pt}
  \caption{Information and model configurations on different datasets.}
  \label{Tabel_ModelConfig}
  \centering
  \begin{tabular}{l c c c c c}
    \toprule
    Dataset             & MNIST-ONE & MNIST-TWO & CIFAR-10 & CUB-200  \\
    \midrule	        
    Image Size        & 32               & 64              & 32                  & 64   \\
    \#Images           & 60,000             & 60,000             & 50,000             & 5,994 \\
    \#Timesteps       & 2                  & 3                   & 2             & 2                            \\
	\#Parameters    & 5.25M/4.11M     & 7.53M/6.33M          & 5.26M/4.11M     & 27.3M/6.34M                       \\	
   \bottomrule
  \end{tabular}
\end{table}

\subsection{Results on MNIST-ONE}

\label{Appendix_MNIST-ONE}

\begin{figure}[b]
\begin{minipage}{0.5\linewidth}
\center
\includegraphics[scale=0.12]{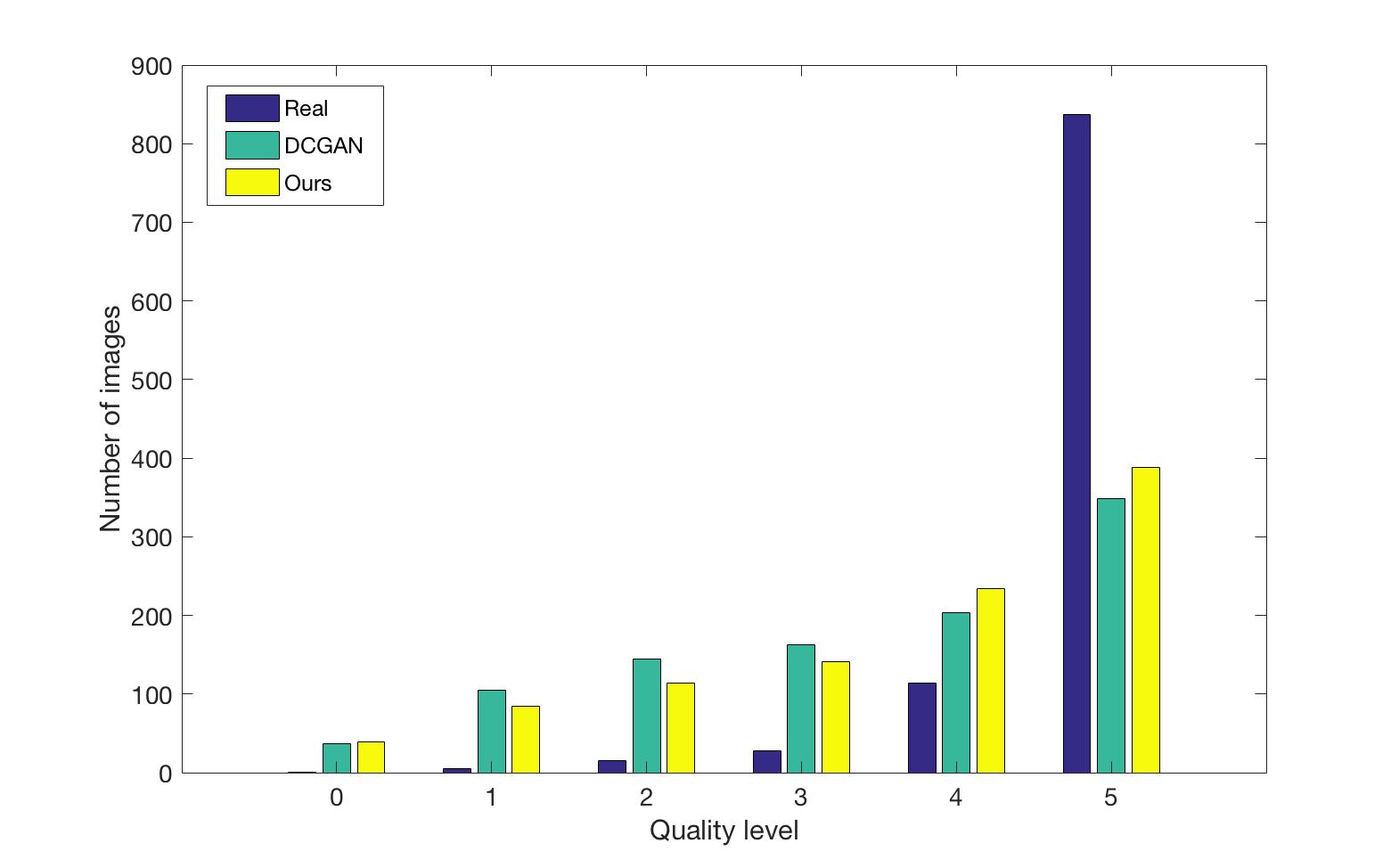}
\end{minipage}
\begin{minipage}{0.5\linewidth}
\center
\includegraphics[scale=0.24]{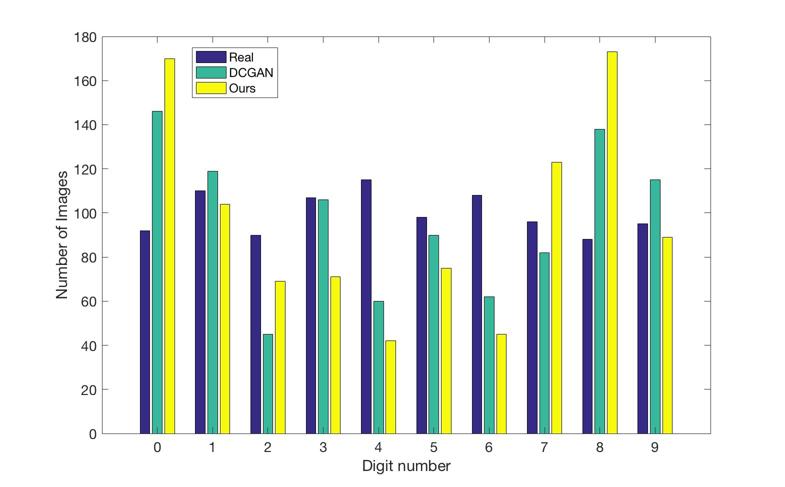}
\end{minipage}
\caption{Statistics of annotations in human studies on MNIST-ONE. Left: distribution of quality level; Right: distribution of recognized digit categories.}
\label{Fig_MNISTONE_Stat}
\end{figure}

\begin{figure}[h]
\begin{minipage}{0.16\linewidth}
\center
\includegraphics[scale=0.33]{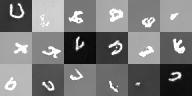}
\end{minipage}
\begin{minipage}{0.16\linewidth}
\center
\includegraphics[scale=0.33]{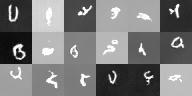}
\end{minipage}
\begin{minipage}{0.16\linewidth}
\center
\includegraphics[scale=0.33]{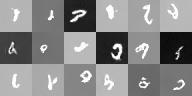}
\end{minipage}
\begin{minipage}{0.16\linewidth}
\center
\includegraphics[scale=0.33]{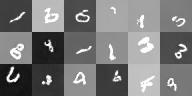}
\end{minipage}
\begin{minipage}{0.16\linewidth}
\center
\includegraphics[scale=0.33]{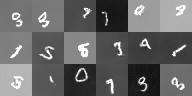}
\end{minipage}
\begin{minipage}{0.16\linewidth}
\center
\includegraphics[scale=0.33]{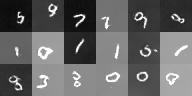}
\end{minipage}

\begin{minipage}{0.16\linewidth}
\center
\includegraphics[scale=0.33]{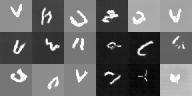}
\end{minipage}
\begin{minipage}{0.16\linewidth}
\center
\includegraphics[scale=0.33]{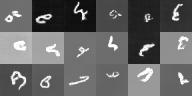}
\end{minipage}
\begin{minipage}{0.16\linewidth}
\center
\includegraphics[scale=0.33]{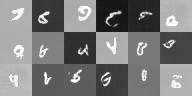}
\end{minipage}
\begin{minipage}{0.16\linewidth}
\center
\includegraphics[scale=0.33]{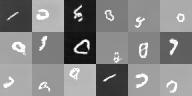}
\end{minipage}
\begin{minipage}{0.16\linewidth}
\center
\includegraphics[scale=0.33]{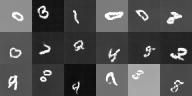}
\end{minipage}
\begin{minipage}{0.16\linewidth}
\center
\includegraphics[scale=0.33]{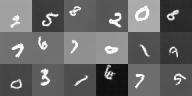}
\end{minipage}
\caption{Qualitative comparison on MNIST-ONE. Top three rows are samples generated by DCGAN. Bottom three rows are samples generated by LR-GAN. The quality level increases from left to right as determined via human studies.}
\label{Fig_MNISTComparison}
\end{figure}

We conduct human studies on generation results on MNIST-ONE. Specifically, we generate 1,000 images using both LR-GAN and DCGAN. As references, we also include 1000 real images. Then we ask the users on AMT to label each image to be one of the digits (0-9). We also provide them an option `non recognizable' in case the generated image does not seem to contain a digit. Each image was judged by 5 unique workers. Similar to CIFAR-10, if an image is recognized to be the same digit by all 5 users, it is assigned to quality level 5. If it is not recognizable according to all users, it is assigned to quality level 0. Fig.~\ref{Fig_MNISTONE_Stat} (left) shows the number of images assigned to all six quality levels. Compared to DCGAN, our model generated more samples with high quality levels. As expected, the real images have many samples with high quality levels. In Fig.~\ref{Fig_MNISTONE_Stat} (right), we show the number of images that are recognized to each digit category (0-9). For qualitative comparison, we show examplar images at each quality level in Fig.~\ref{Fig_MNISTComparison}. From left to right, the quality level increases from 0 to 5. As expected, the images with higher quality level are more clear.

\textcolor{black}{For quantitative evaluation, we use the same way as for CIFAR-10. The classifier model used for contextual Inception Score is trained based on the training set. We generate 60,000 samples based on DCGAN and LR-GAN for evaluation, respectively. To obtain the Adversarial Accuracy and Adversarial Divergence, we first train 10 generators for 10 digit categories separately, and then use the generated samples to train the classifier. As shown in Table~\ref{Tabel_QuantMNISTONE}, our model has higher scores than DCGAN on both standard and contextual Inception Score. Also, our model has a slightly higher adversarial accuracy, and lower adversarial divergence than DCGAN. We find that the all three image sets have low standard Inception Scores. This is mainly because the Inception net is trained on ImageNet, which has a very different data distribution from the MNIST dataset. Based on this, we argue that the standard Inception Score is not suitable for some image datasets.}

\begin{table}[t] \footnotesize
\setlength{\tabcolsep}{5.5pt}
  \caption{Quantitative comparison on MNIST-ONE.}
  \label{Tabel_QuantMNISTONE}
  \centering
  \begin{tabular}{l c c c}
    \toprule
    Training Data           & Real Images &  DCGAN & Ours  \\
    \midrule	    
     Inception Score\textsuperscript{\textdagger}             &    1.83$\pm$0.01              &2.03$\pm$0.01         &  2.06$\pm$0.01 \\
     Inception Score\textsuperscript{\textdagger\textdagger}             &    9.15$\pm$0.04              & 6.42$\pm$0.03         &  7.15$\pm$0.04 \\         
     Adversarial Accuracy                &     95.22 $\pm$ 0.25             & 26.12 $\pm$ 0.07                    &  26.61 $\pm$ 0.06 \\
	Adversarial Divergence Score               & 0                              &   8.47 $\pm$ 0.03                    & 8.39 $\pm$ 0.04                      \\
   \bottomrule
      \multicolumn{4}{l}{\textsuperscript{\textdagger}\footnotesize{Evaluate using the pre-trained Inception net as \cite{ImprovedGAN}}} \\
   \multicolumn{4}{l}{\textsuperscript{\textdagger\textdagger}\footnotesize{Evaluate using the supervisedly trained classifier based on the discriminator in LR-GAN.}}	
  \end{tabular}
\end{table}

\subsection{More results on CUB-200}
\label{Appendix_CUB200}
\begin{figure}[!ht]
\begin{minipage}{0.138\linewidth}
\center
\includegraphics[scale=0.17]{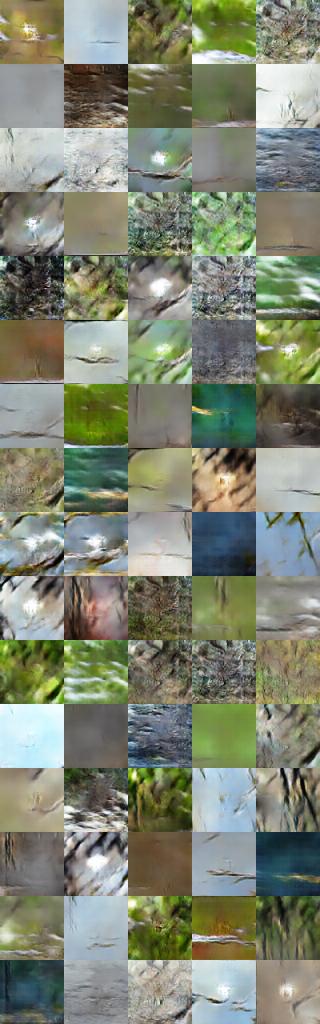}
\end{minipage}
\begin{minipage}{0.138\linewidth}
\center
\includegraphics[scale=0.17]{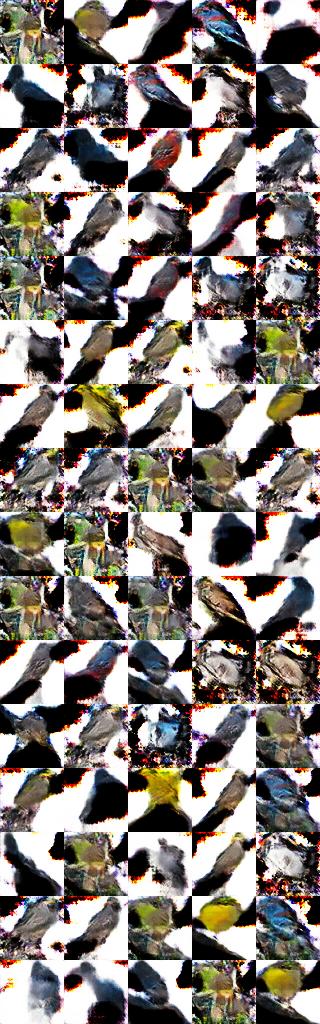}
\end{minipage}
\begin{minipage}{0.138\linewidth}
\center
\includegraphics[scale=0.17]{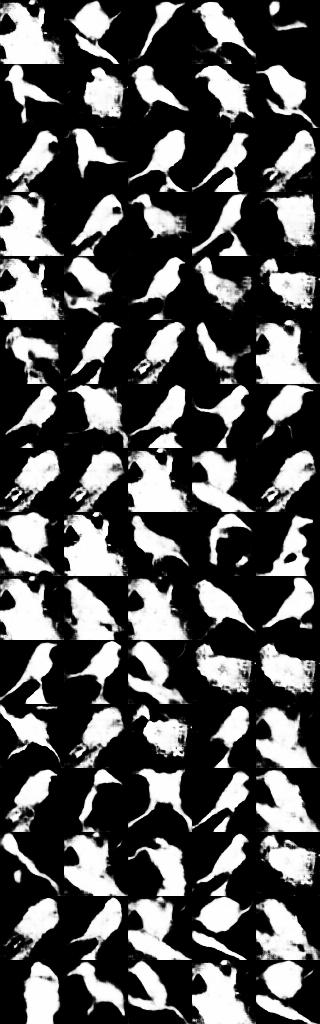}
\end{minipage}
\begin{minipage}{0.138\linewidth}
\center
\includegraphics[scale=0.17]{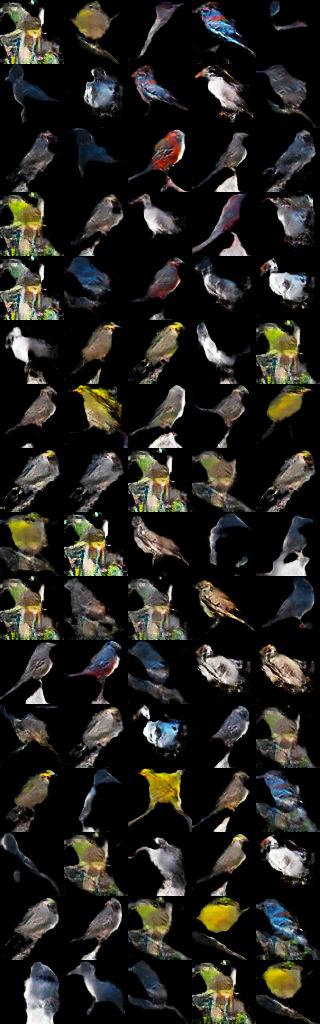}
\end{minipage}
\begin{minipage}{0.138\linewidth}
\center
\includegraphics[scale=0.17]{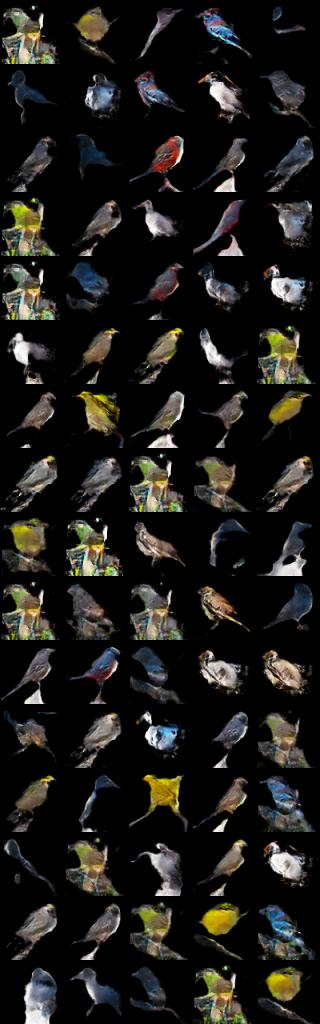}
\end{minipage}
\begin{minipage}{0.138\linewidth}
\center
\includegraphics[scale=0.17]{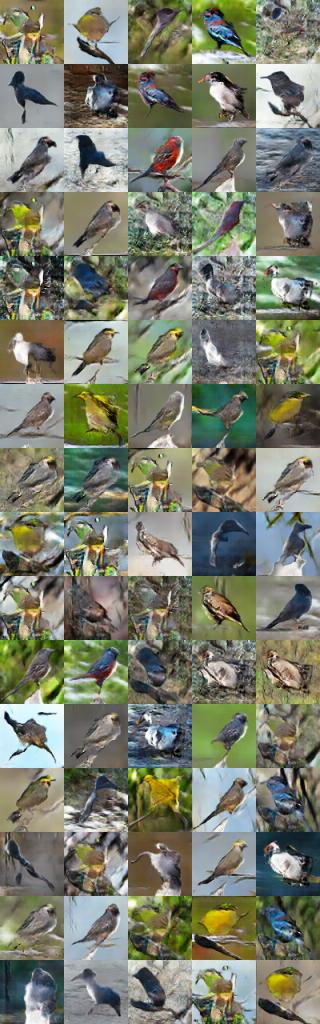}
\end{minipage}
\begin{minipage}{0.138\linewidth}
\center
\includegraphics[scale=0.17]{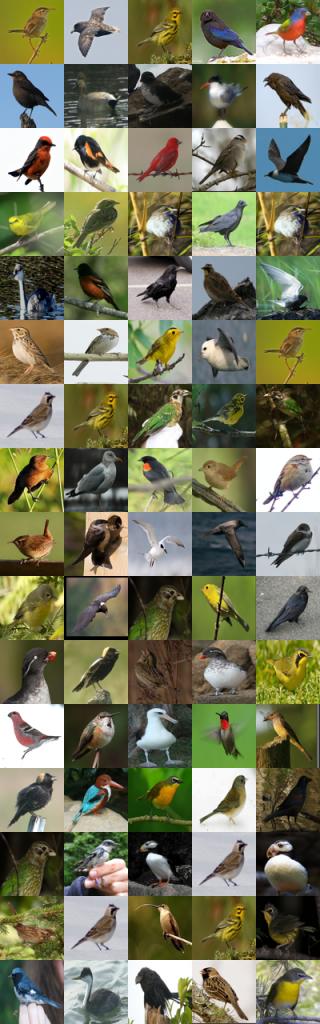}
\end{minipage}
\caption{\textcolor{black}{Generation results of our model on CUB-200 when setting minimal allowed scale to 1.1. From left to right, the blocks show the generated background images, foreground images, foreground masks, foreground images carved out by masks, carved foreground images after spatial transformation. The sixth and seventh blocks are final composite images and the nearest neighbor real images.}}
\label{Fig_CUBOutputs_1.1}
\end{figure}

In this experiment, we reduce the minimal allowed object scale to 1.1, which allows the model to generate larger foreground objects. The results are shown in Fig.~\ref{Fig_CUBOutputs_1.1}. Similar to the results when the constraint is 1.2, the crisp bird-like masks are generated automatically by our model.

\subsection{More results on CIFAR-10}
\subsubsection{Qualitative Results}
\begin{figure}[h]
\begin{minipage}{0.138\linewidth}
\center
\includegraphics[scale=0.33]{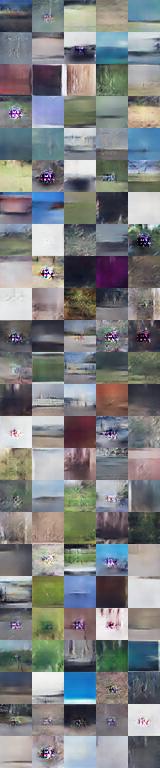}
\end{minipage}
\begin{minipage}{0.138\linewidth}
\center
\includegraphics[scale=0.33]{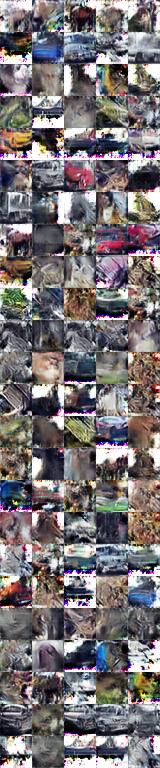}
\end{minipage}
\begin{minipage}{0.138\linewidth}
\center
\includegraphics[scale=0.33]{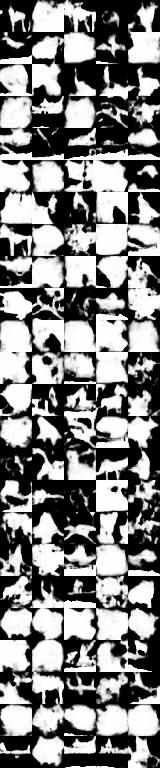}
\end{minipage}
\begin{minipage}{0.138\linewidth}
\center
\includegraphics[scale=0.33]{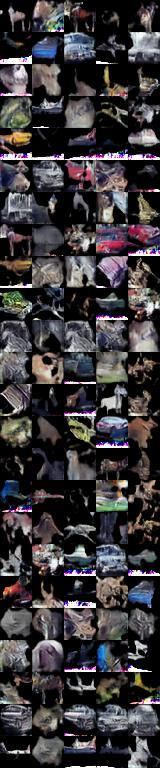}
\end{minipage}
\begin{minipage}{0.138\linewidth}
\center
\includegraphics[scale=0.33]{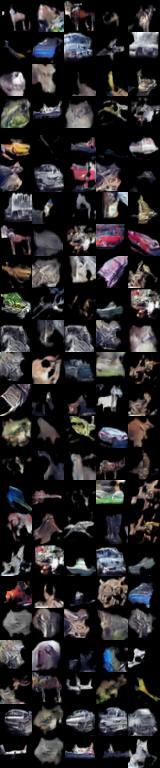}
\end{minipage}
\begin{minipage}{0.138\linewidth}
\center
\includegraphics[scale=0.33]{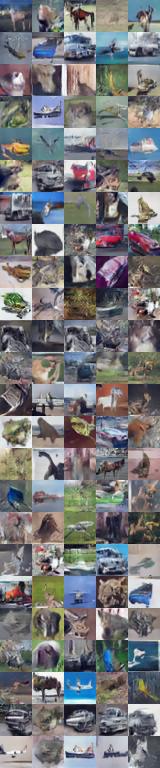}
\end{minipage}
\begin{minipage}{0.138\linewidth}
\center
\includegraphics[scale=0.33]{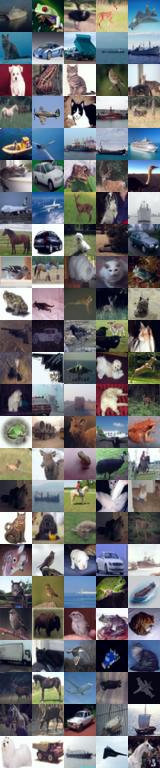}
\end{minipage}
\caption{\textcolor{black}{Generation results of our model on CIFAR-10 with minimal allowed scale be 1.1, From left to right, the layout is same to Fig.~\ref{Fig_CUBOutputs_1.1}.}}
\label{Fig_CIFAROutputs_1.1}
\end{figure}

In Fig.~\ref{Fig_CIFAROutputs_1.1}, we show more results on CIFAR-10 when setting minimal allowed object scale to 1.1. The rightmost column block also shows the training images that are closest to the generated images (cosine similarity in pixel space). We can see our model does not memorize the training data.

\begin{figure}[t]
\begin{minipage}{1\linewidth}
\center
\includegraphics[scale=0.335]{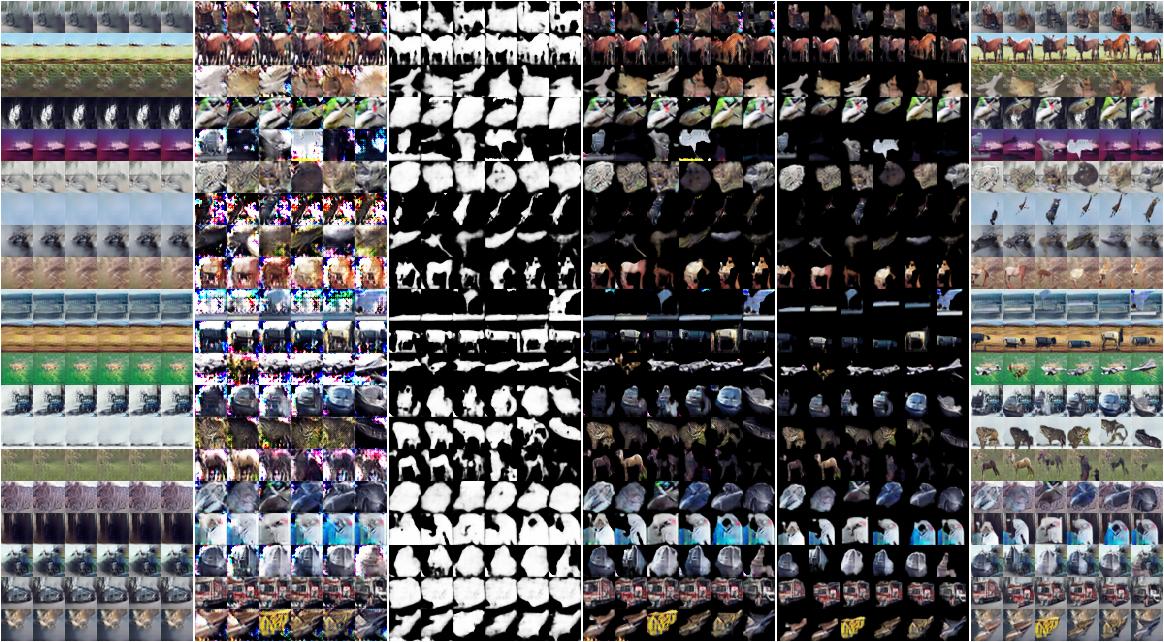}
\end{minipage}
\caption{Walking in the latent foreground space by fixing backgrounds in our model on CIFAR-10. From left to right, the blocks are: generated background images, foreground images, foreground masks, foreground images carved out by masks, carved out foreground images after spatial transformation, and final composite images. Each row has the same background, but different foregrounds.}
\label{Fig_CIFAROutputs_fixbg}
\end{figure} 
%
%

\subsubsection{Walking in the latent space}
Similar to DCGAN, we also show results by walking in the latent space. Note that our model has two or more inputs. So we can walk along any of them or their combination. 
In Fig.~\ref{Fig_CIFAROutputs_fixbg}, we generate multiple foregrounds for the same fixed generated background. We find that our model consistently generates contextually compatible foregrounds. 
For example, for the grass-like backgrounds, the foreground generator generates horses and deer, and airplane-like objects for the blue sky. 

\subsubsection{Word cloud based on human study}
\label{Appendix_CIFAR_FIxedBG}
\begin{figure}[h]
  \begin{minipage}{0.33\textwidth}
    \center
\includegraphics[scale=0.14]{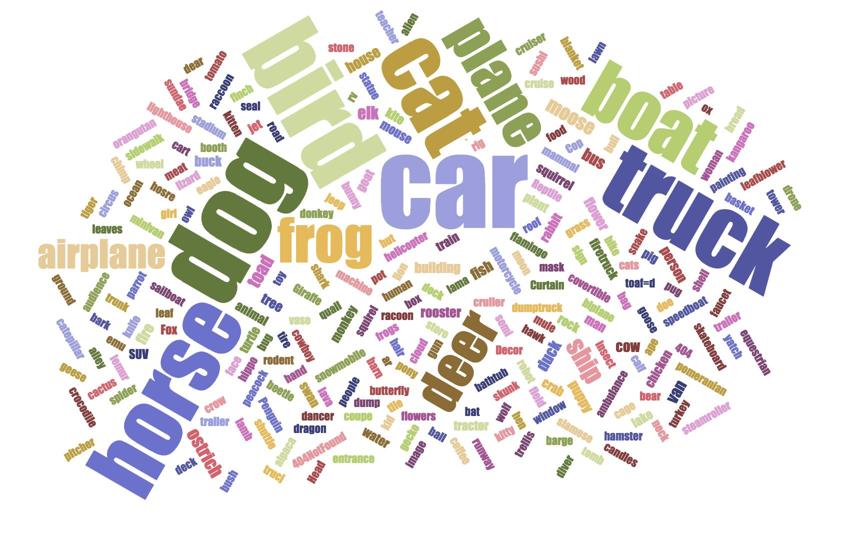}
  \end{minipage}   
  \begin{minipage}{0.33\textwidth}
    \center
\includegraphics[scale=0.14]{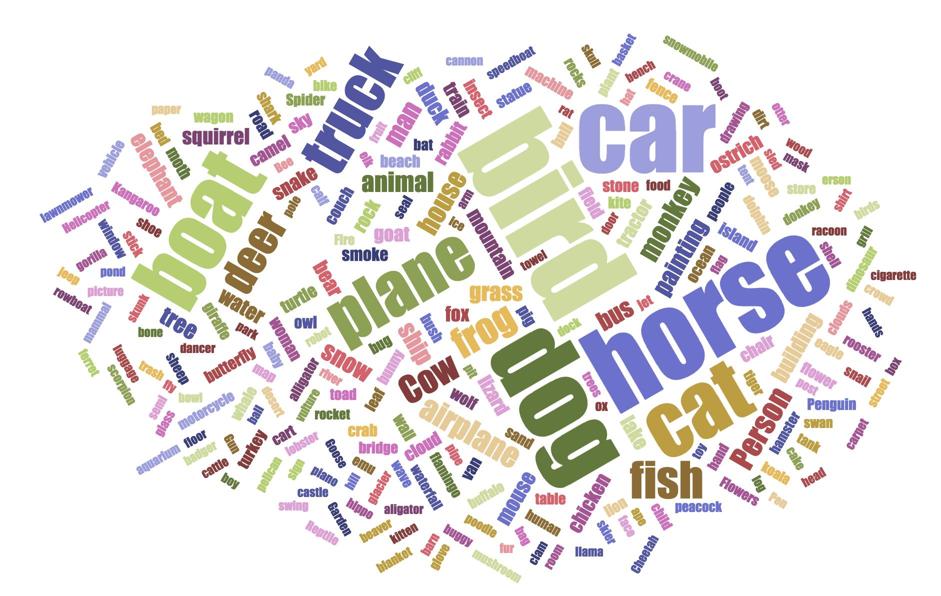}
  \end{minipage}    
  \begin{minipage}{0.33\textwidth}
    \center
\includegraphics[scale=0.15]{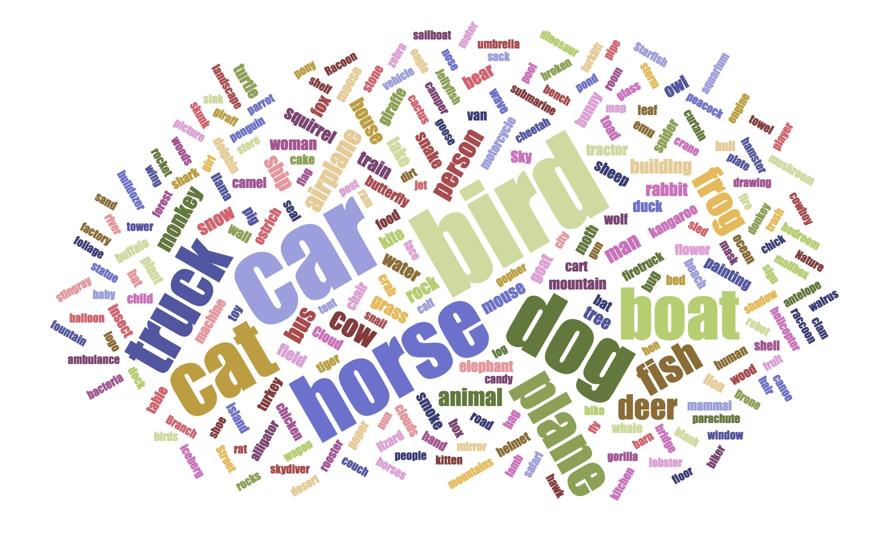}
  \end{minipage}      
\caption{Statistics of annotations in human studies on CIFAR-10. Left to right: word cloud for real images, images generated by DCGAN, images generated by LR-GAN.}
\label{Fig_WordCloud}
\end{figure}
As we mentioned above, we conducted human studies on CIFAR-10. Besides asking people  to select a name from a list for an image, we also conducted another human study where we ask people to use one word (free-form) to describe the main object in the image. Each image was `named' by 5 unique people. We generate word clouds for real images, images generated by DCGAN and LR-GAN, as shown in Fig.~\ref{Fig_WordCloud}. 

\subsection{Results on LFW face dataset}

\begin{figure}[!ht]
\begin{minipage}{1\linewidth}
\center
\includegraphics[scale=0.49]{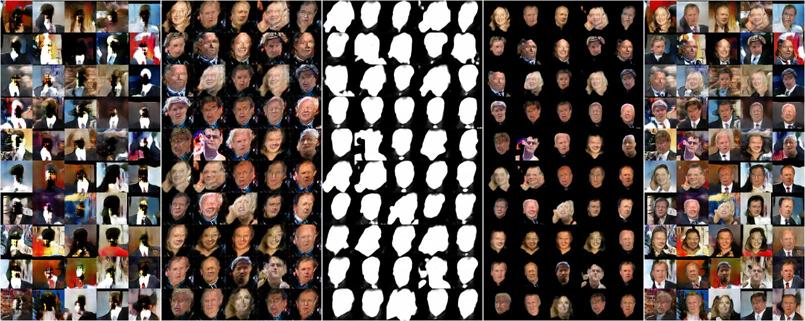}
\end{minipage}
\caption{Generation results of our model on LFW. From left to right, the blocks are: generated background images, foreground images, foreground masks, carved out foreground images after spatial transformation, and final composite images.}
\label{Fig_LFWOutputs}
\end{figure} 

We conduct experiment on face images in LFW dataset \citep{LFWTech}. Different from previous works which work on cropped and aligned faces, we directly generate the original images which contains a large portion of backgrounds. This configuration helps to verify the efficiency of LR-GAN to model the object appearance, shape and pose. In Fig.~\ref{Fig_LFWOutputs}, we show the (intermediate) generation results of LR-GAN. Surprisingly, without any supervisions, the model generated background and faces in separate steps, and the generated masks accurately depict face shapes. Moreover, the model learns where to place the generated faces so that the whole image looks natural. For comparison, please refer to \citep{CompositeGAN} which does not model the transformation. We can find the generation results degrade much.

\subsection{Statistics on Transformation Matrices}

\begin{figure}[!ht]
\begin{minipage}{0.246\linewidth}
\center
\includegraphics[scale=0.13]{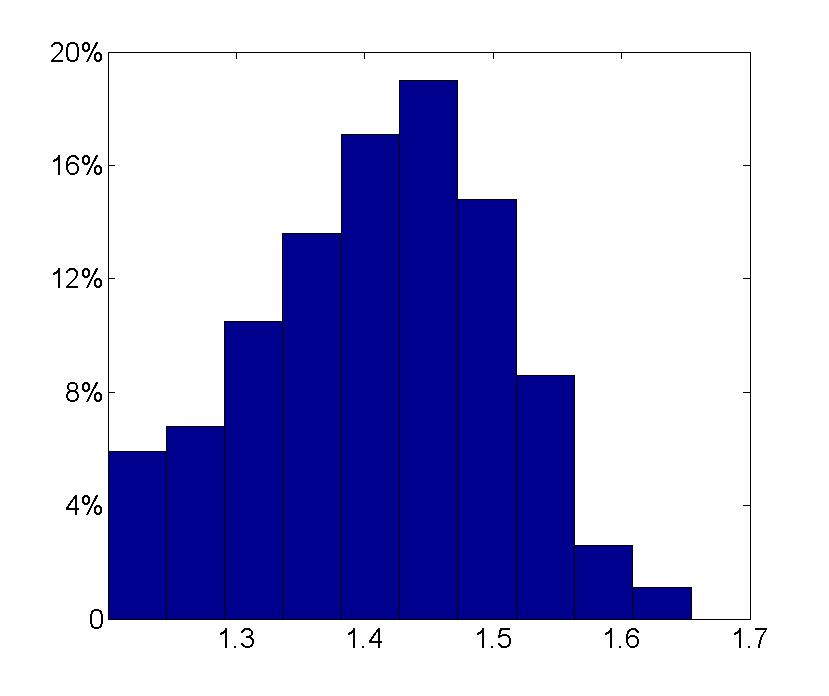}
\end{minipage}
\begin{minipage}{0.246\linewidth}
\center
\includegraphics[scale=0.13]{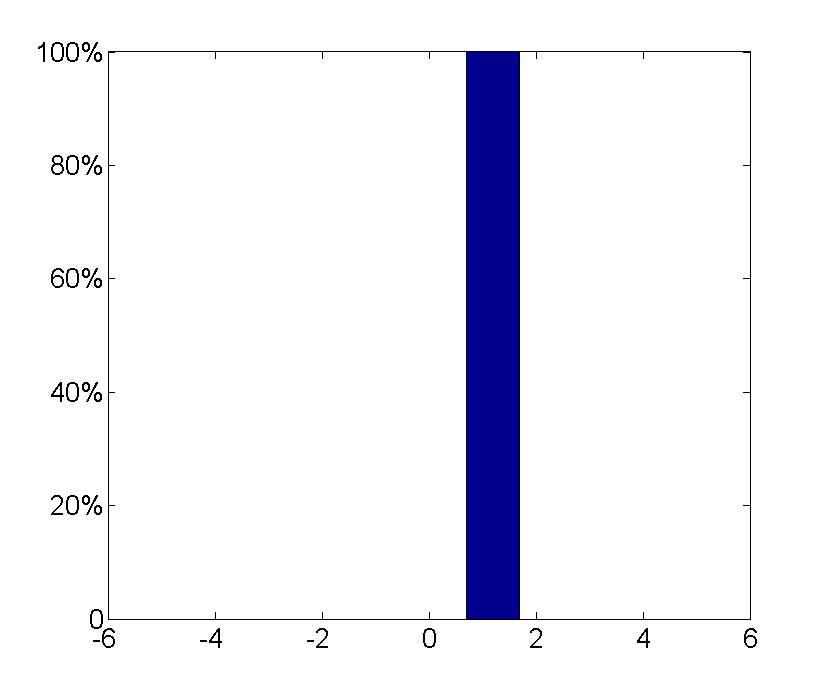}
\end{minipage}
\begin{minipage}{0.246\linewidth}
\center
\includegraphics[scale=0.13]{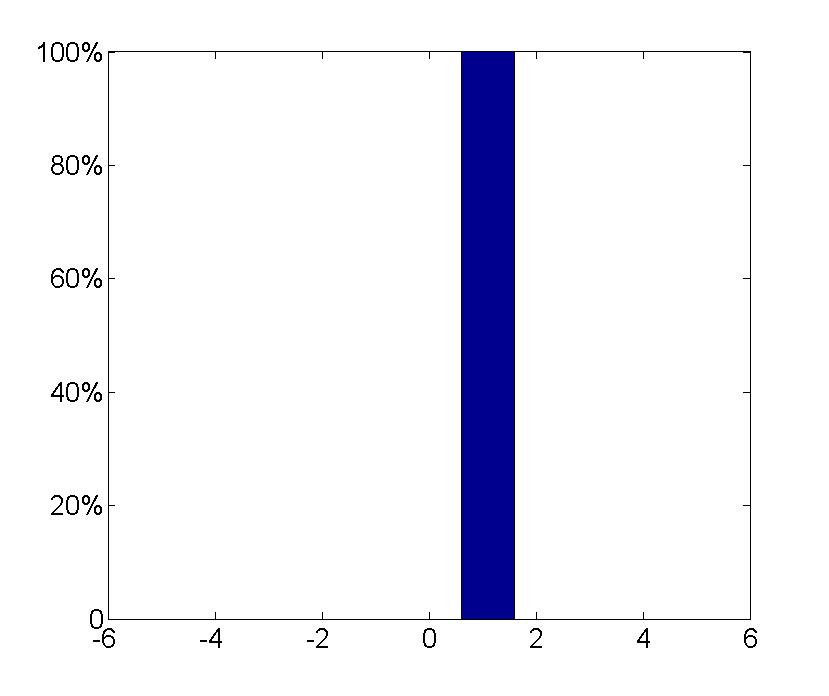}
\end{minipage}
\begin{minipage}{0.246\linewidth}
\center
\includegraphics[scale=0.13]{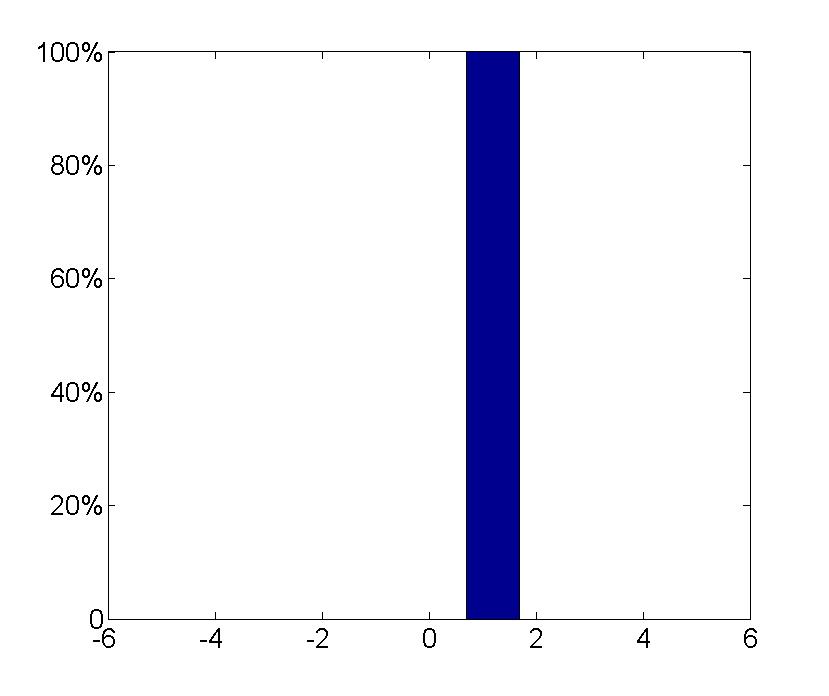}
\end{minipage}

\begin{minipage}{0.246\linewidth}
\center
\includegraphics[scale=0.13]{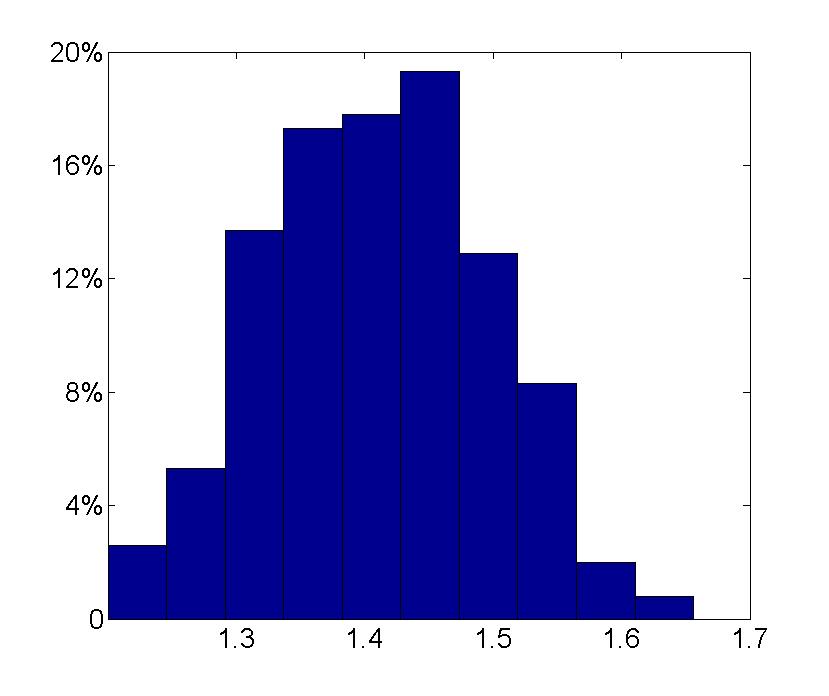}
\end{minipage}
\begin{minipage}{0.246\linewidth}
\center
\includegraphics[scale=0.13]{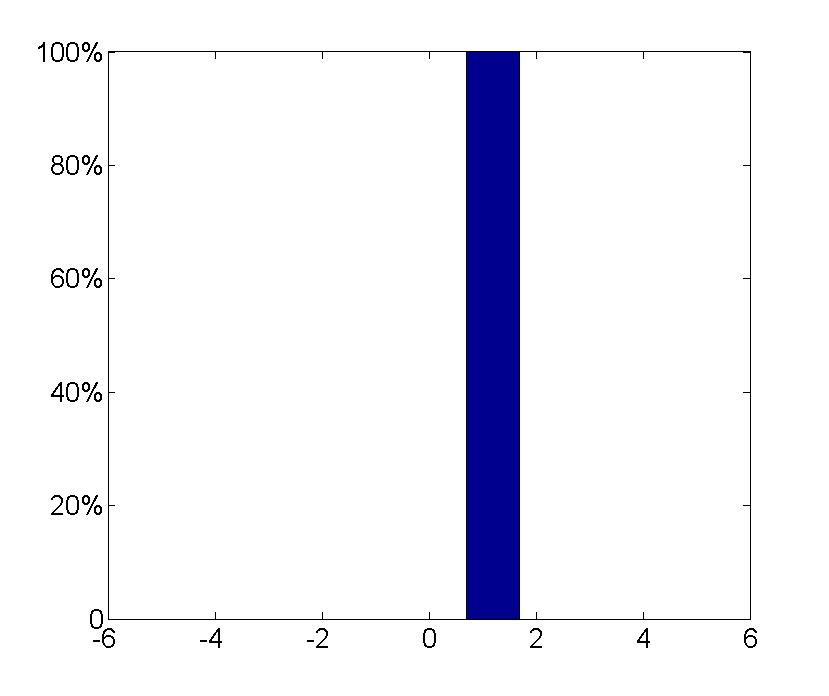}
\end{minipage}
\begin{minipage}{0.246\linewidth}
\center
\includegraphics[scale=0.13]{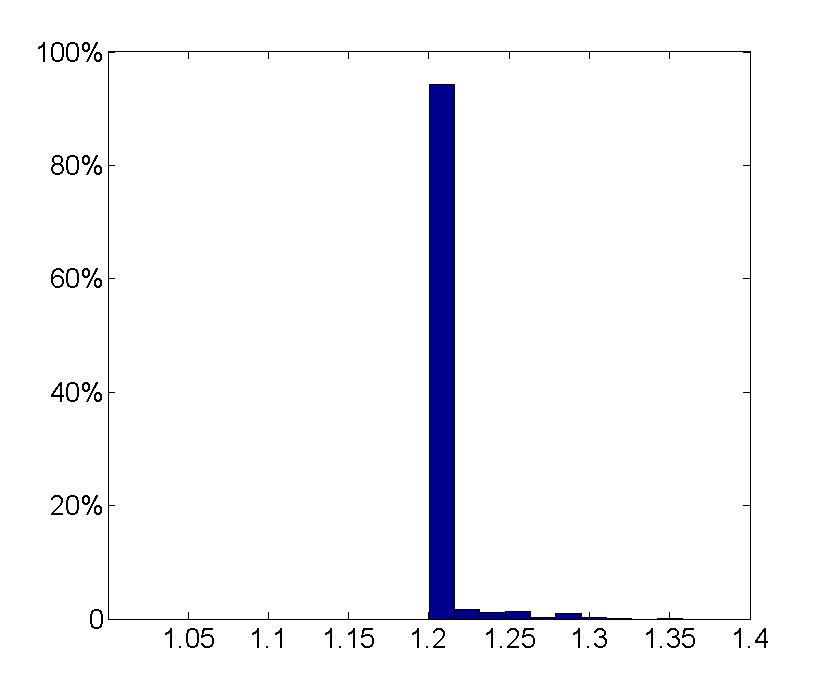}
\end{minipage}
\begin{minipage}{0.246\linewidth}
\center
\includegraphics[scale=0.13]{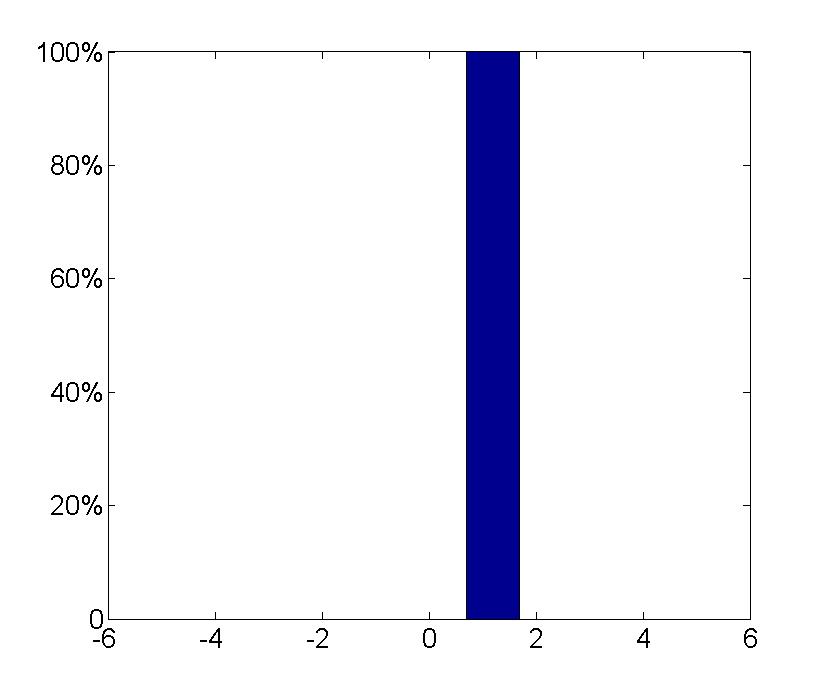}
\end{minipage}

\begin{minipage}{0.246\linewidth}
\center
\includegraphics[scale=0.13]{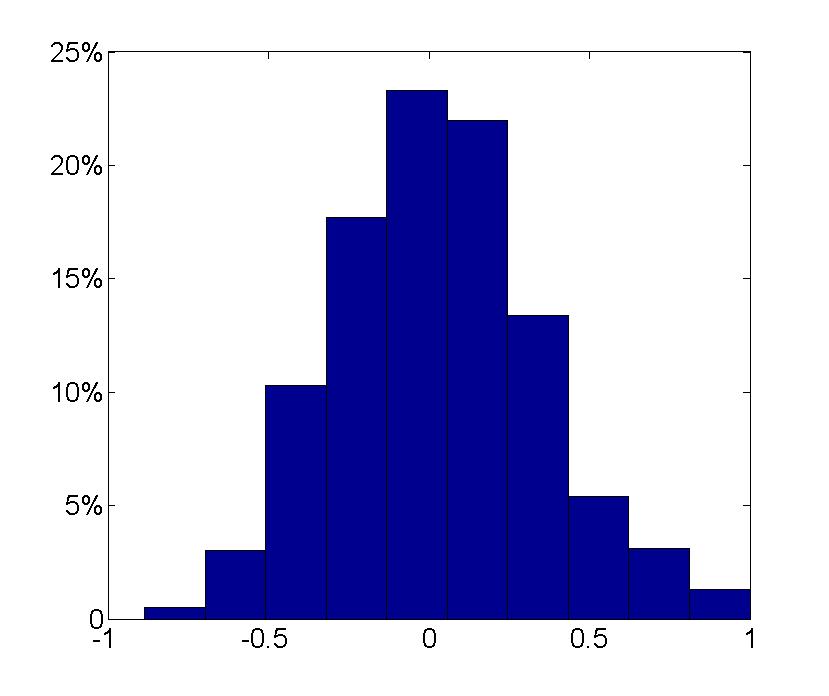}
\end{minipage}
\begin{minipage}{0.246\linewidth}
\center
\includegraphics[scale=0.13]{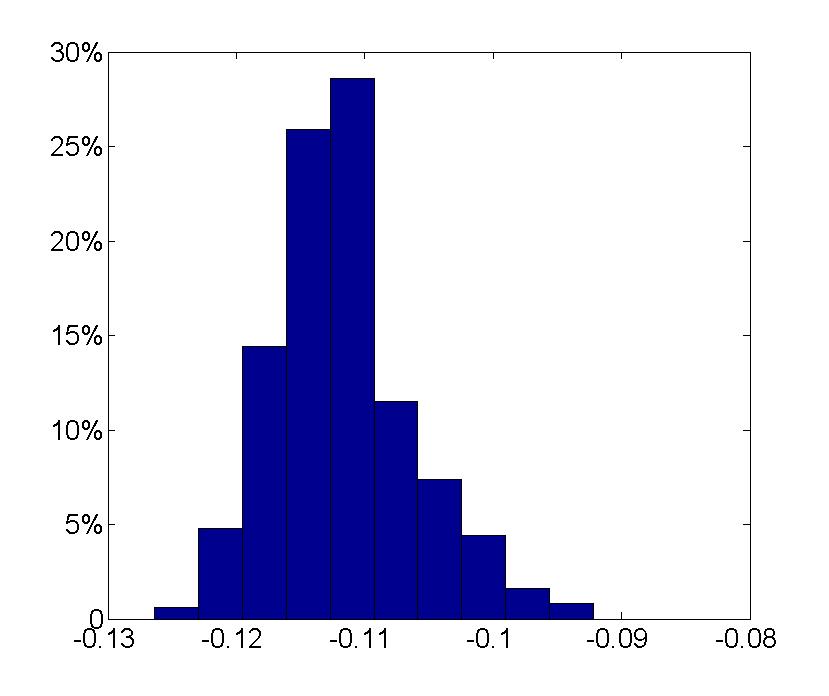}
\end{minipage}
\begin{minipage}{0.246\linewidth}
\center
\includegraphics[scale=0.13]{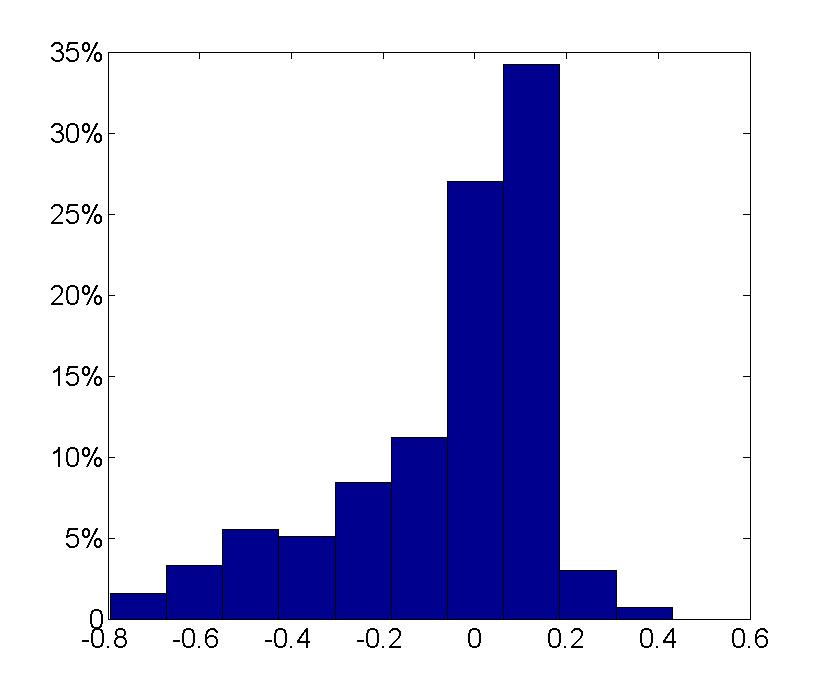}
\end{minipage}
\begin{minipage}{0.246\linewidth}
\center
\includegraphics[scale=0.13]{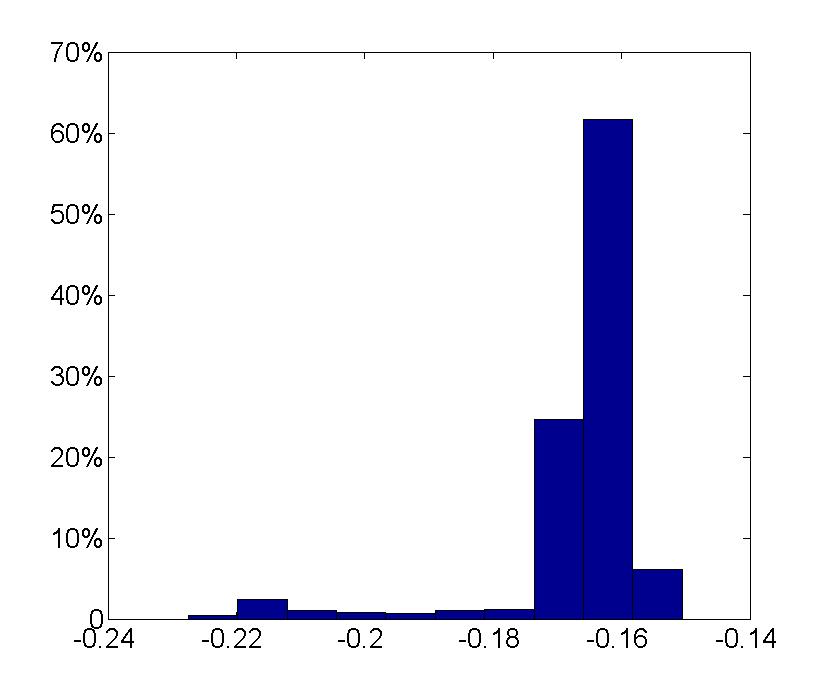}
\end{minipage}

\begin{minipage}{0.246\linewidth}
\center
\includegraphics[scale=0.13]{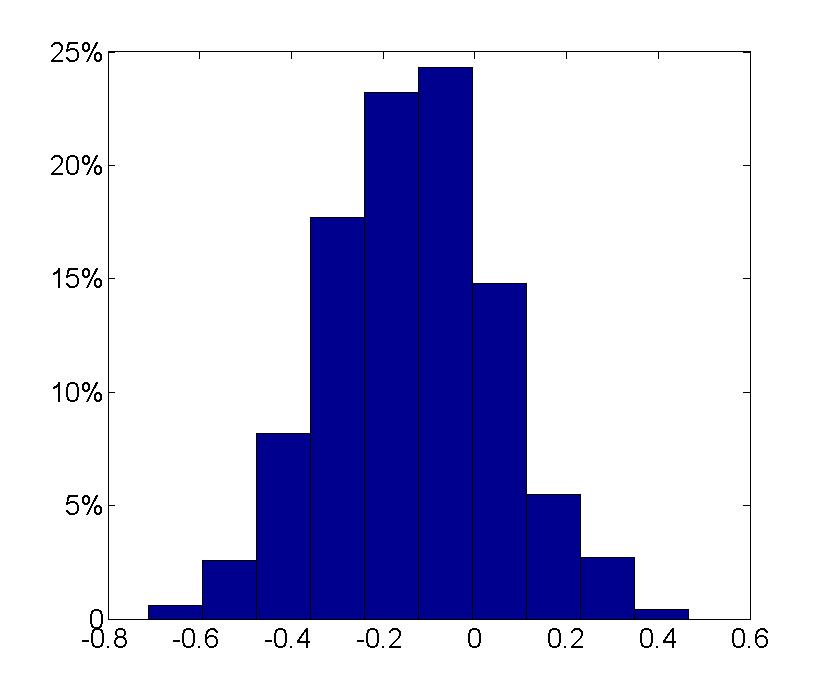}
\end{minipage}
\begin{minipage}{0.246\linewidth}
\center
\includegraphics[scale=0.13]{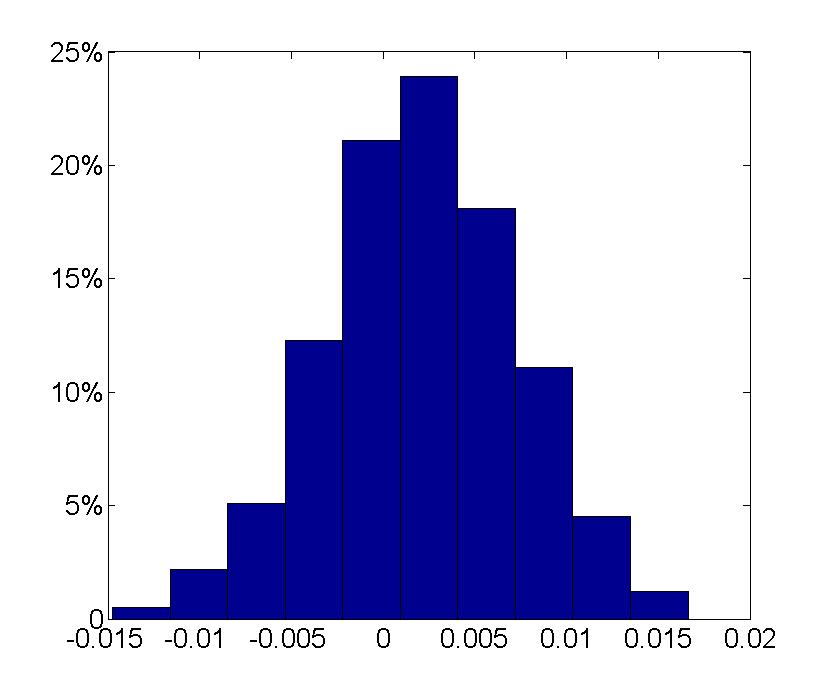}
\end{minipage}
\begin{minipage}{0.246\linewidth}
\center
\includegraphics[scale=0.13]{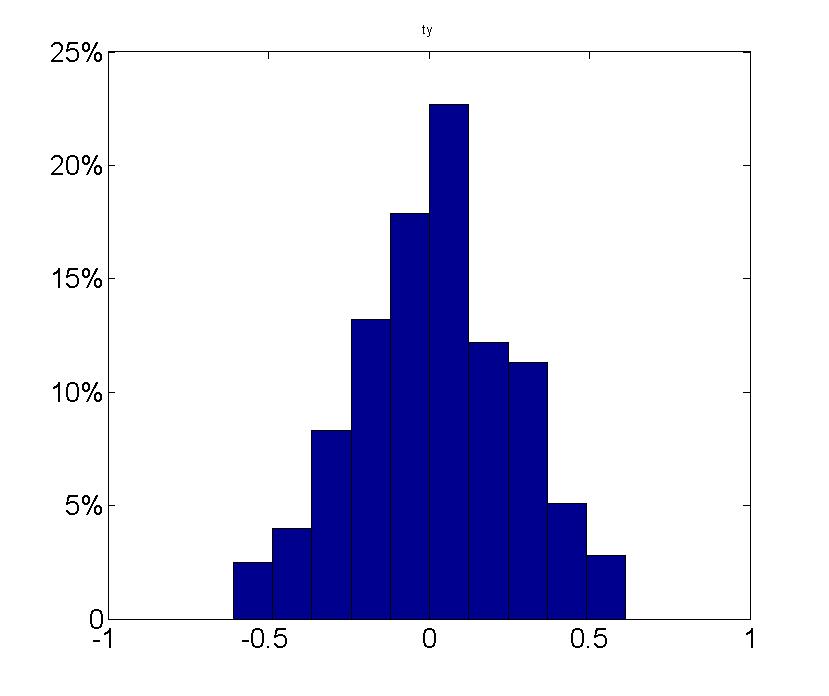}
\end{minipage}
\begin{minipage}{0.246\linewidth}
\center
\includegraphics[scale=0.13]{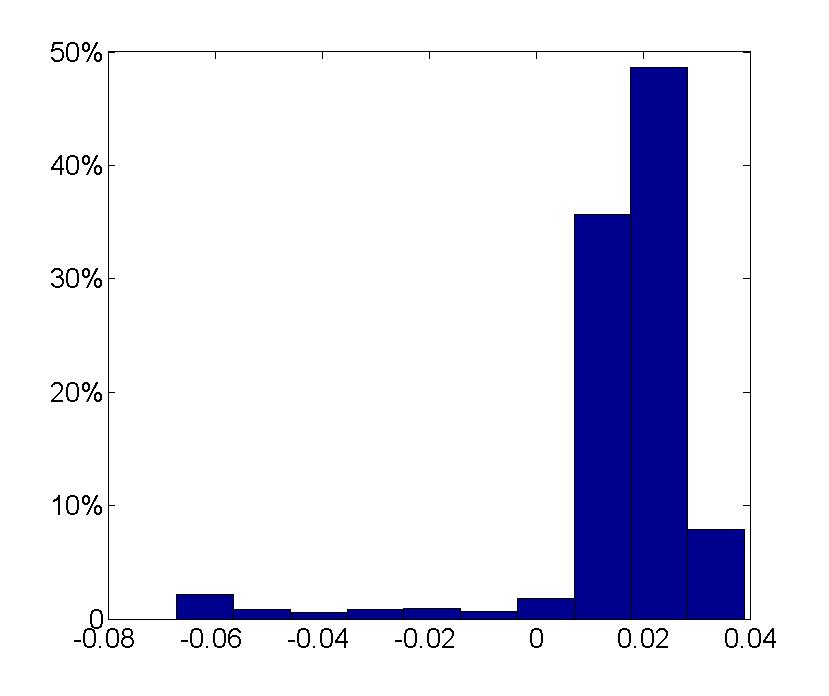}
\end{minipage}

\begin{minipage}{0.246\linewidth}
\center
\includegraphics[scale=0.13]{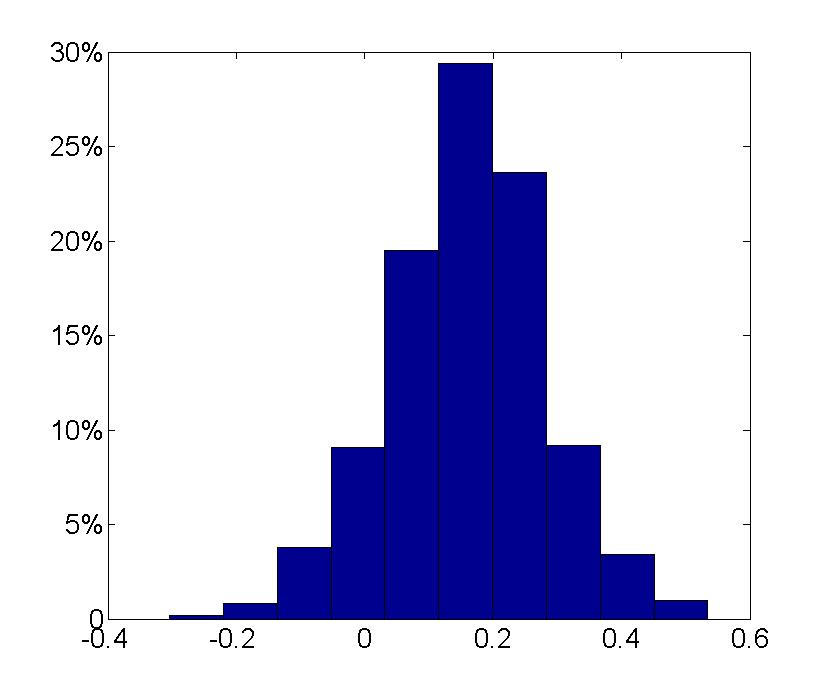}
\end{minipage}
\begin{minipage}{0.246\linewidth}
\center
\includegraphics[scale=0.13]{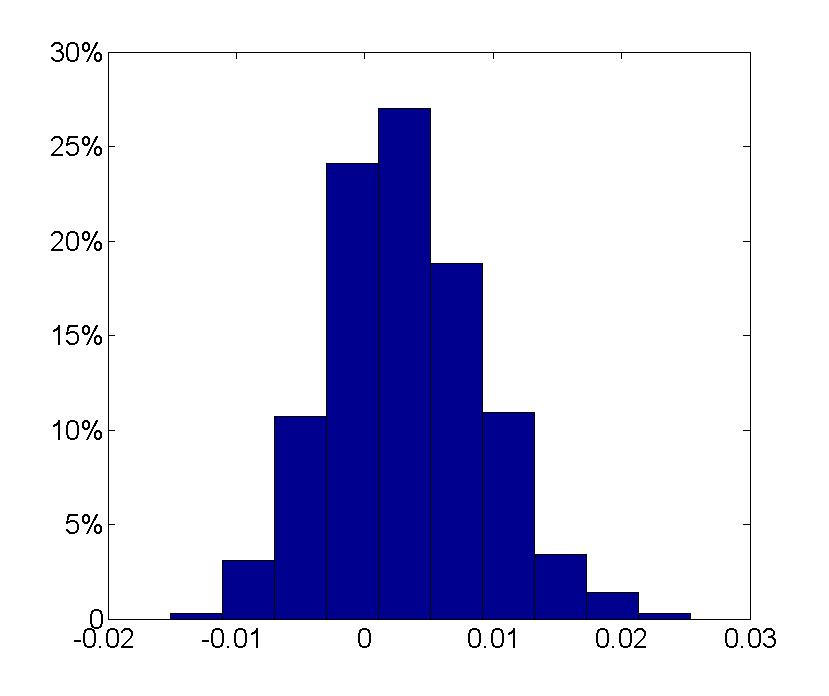}
\end{minipage}
\begin{minipage}{0.246\linewidth}
\center
\includegraphics[scale=0.13]{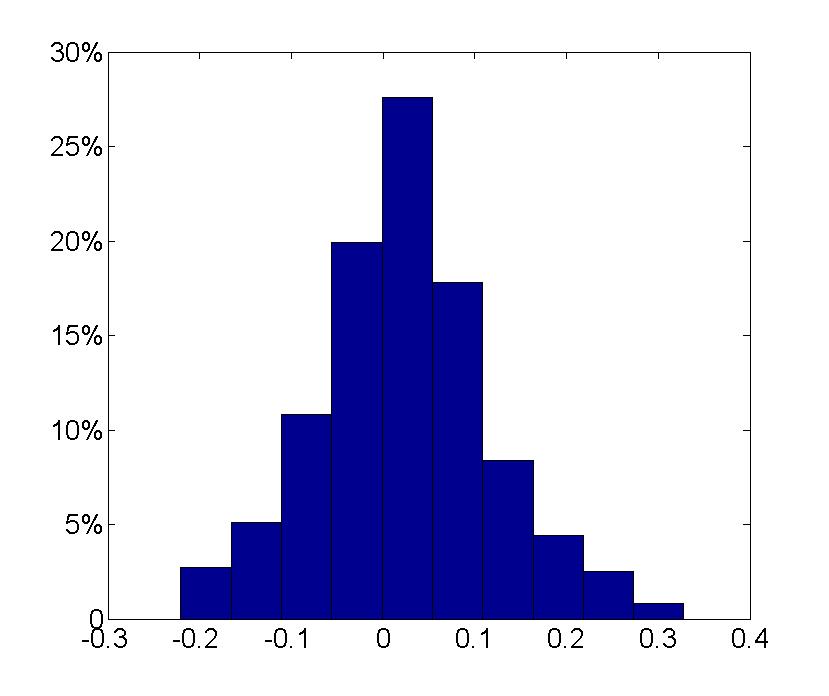}
\end{minipage}
\begin{minipage}{0.246\linewidth}
\center
\includegraphics[scale=0.13]{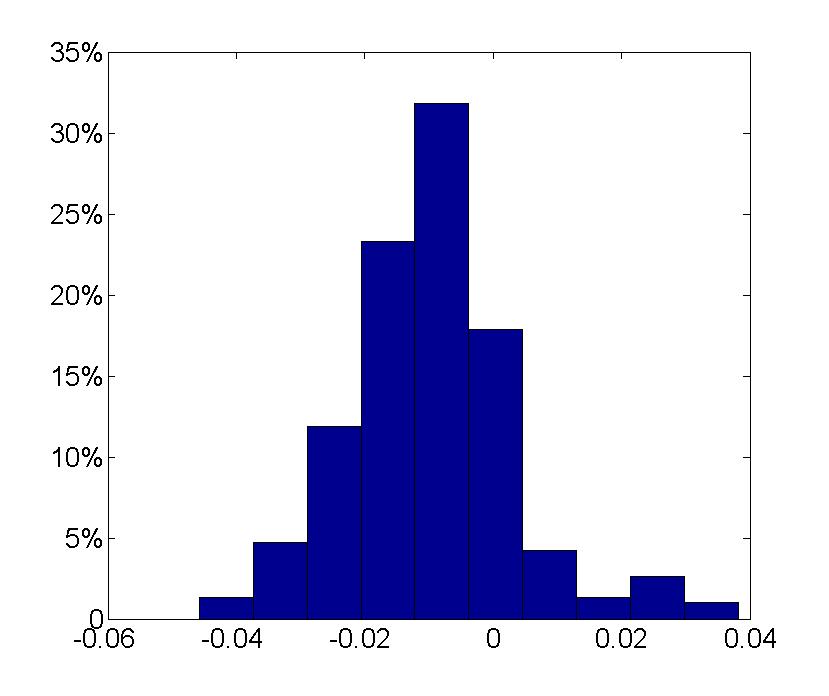}
\end{minipage}

\begin{minipage}{0.246\linewidth}
\center
\includegraphics[scale=0.13]{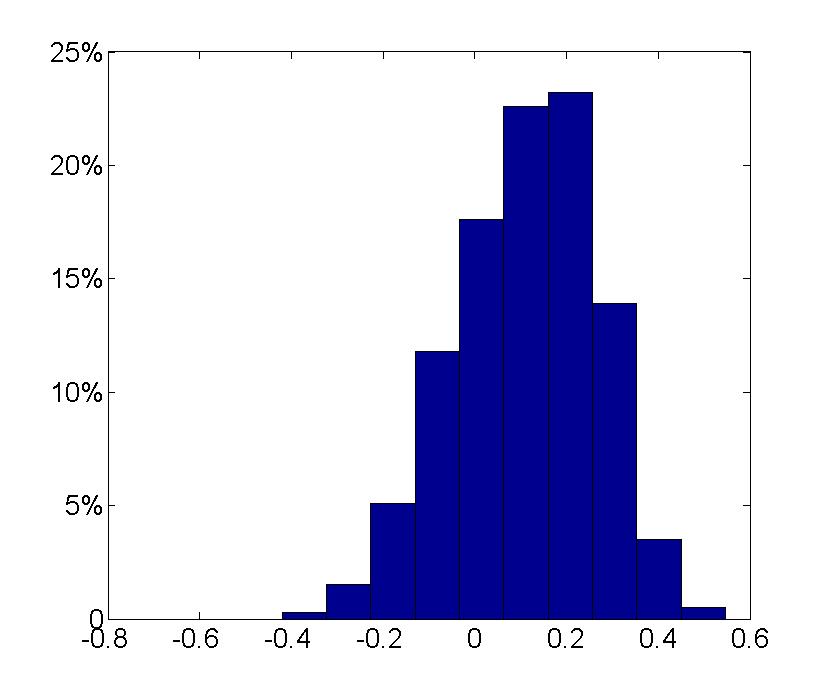}
\end{minipage}
\begin{minipage}{0.246\linewidth}
\center
\includegraphics[scale=0.13]{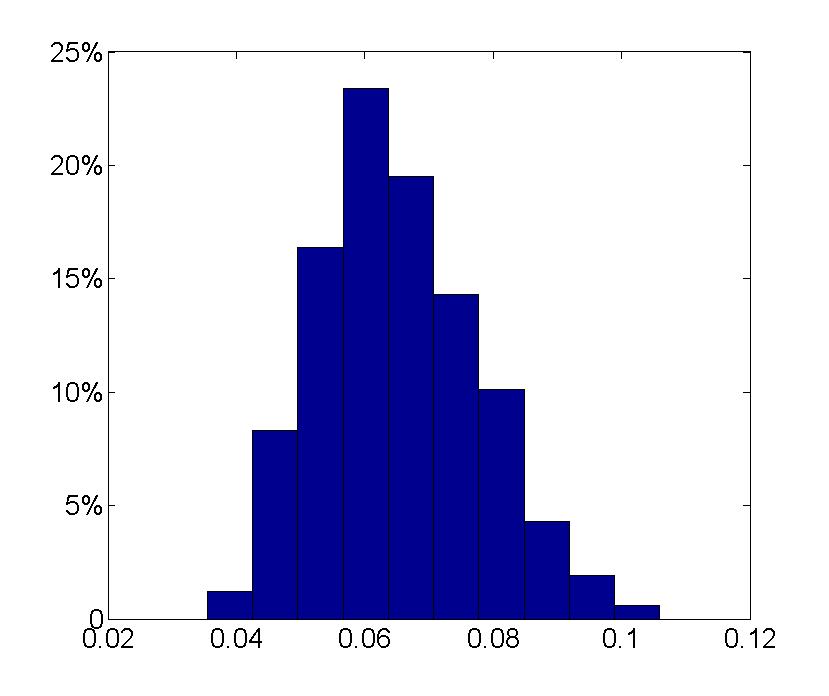}
\end{minipage}
\begin{minipage}{0.246\linewidth}
\center
\includegraphics[scale=0.13]{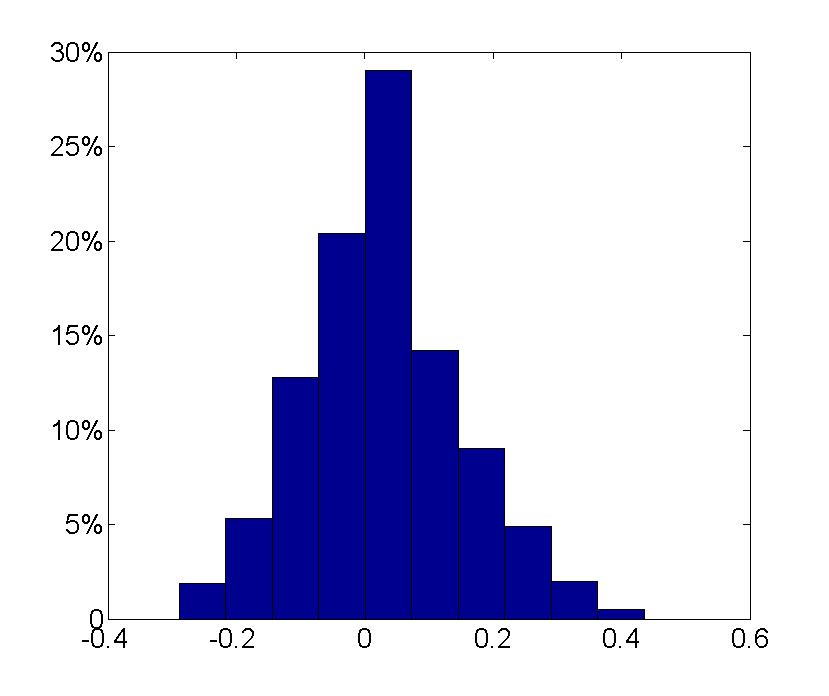}
\end{minipage}
\begin{minipage}{0.246\linewidth}
\center
\includegraphics[scale=0.13]{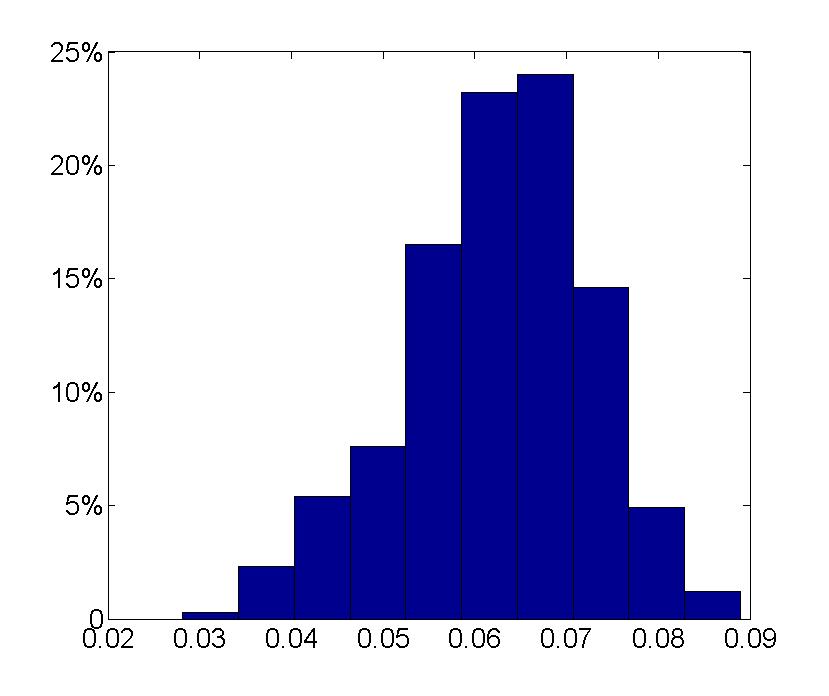}
\end{minipage}
\caption{\textcolor{black}{Histograms of transformation parameters learnt in our model for different datasets. From left to right, the datasets are: MNIST-ONE, CUB-200, CIFAR-10 and LFW. From top to bottom, they are scaling $s_x$, $s_y$, translation $t_x$, $t_y$, and rotation $r_x$, $r_y$ in $x$ and $y$ coordinate, respectively.}}
\label{Fig_Histograms}
\end{figure}

\textcolor{black}{In this part, we analyze the statistics on the transformation matrices generated by our model for different datasets, including MNIST-ONE, CUB-200, CIFAR-10 and LFW. We used affine transformation in our model. So there are 6 parameters, scaling in the x coordinate ($s_x$), scaling in the y coordinate ($s_y$), translation in the x coordinate ($t_x$), translation in the y coordinate ($t_y$), rotation in the x coordinate ($r_x$) and rotation in the y coordinate ($r_y$). In Fig.~\ref{Fig_Histograms}, we show the histograms on different parameters for different datasets.These histograms show that the model produces non-trivial varied scaling, translation and rotation on all datasets. For different datasets, the learned transformation have different patterns. We hypothesize that this is mainly determined by the configurations of objects in the images. For example, on MNIST-ONE, all six parameters have some fluctuations since the synthetic dataset contains digits randomly placed at different locations. For the other three datasets, the scalings converge to single value since the object sizes do not vary much, and the variations on rotation and translation suffice to generate realistic images. Specifically, we can find the generator largely relies on the translation on x coordinate for generating CUB-200. This makes sense since birds in the images have similar scales, orientations but various horizontal locations. For CIFAR-10, since there are 10 different object categories, the configurations are more diverse, hence the generator uses all parameters for generation except for the scaling.  For LFW, since faces have similar configurations, the learned transformations have less fluctuation as well. As a result, we can see that LR-GAN indeed models the transformations on the foreground to generate images.}

\subsection{Conditional Image Generation}
\begin{figure}[!ht]
\begin{minipage}{0.138\linewidth}
\center
\includegraphics[scale=0.33]{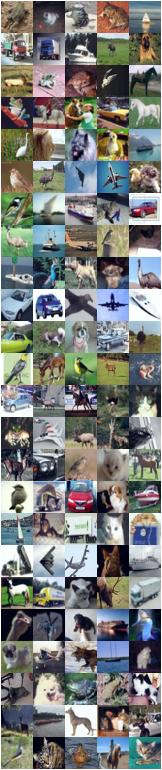}
\end{minipage}
\begin{minipage}{0.138\linewidth}
\center
\includegraphics[scale=0.33]{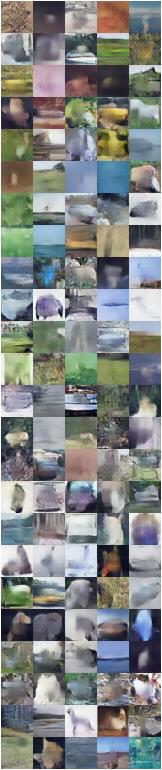}
\end{minipage}
\begin{minipage}{0.138\linewidth}
\center
\includegraphics[scale=0.33]{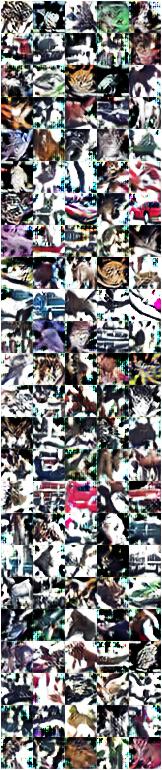}
\end{minipage}
\begin{minipage}{0.138\linewidth}
\center
\includegraphics[scale=0.33]{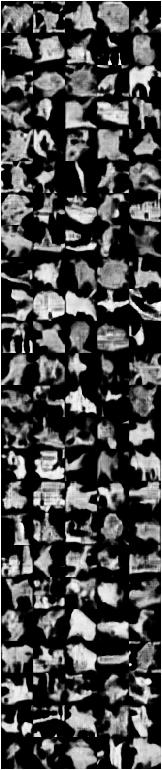}
\end{minipage}
\begin{minipage}{0.138\linewidth}
\center
\includegraphics[scale=0.33]{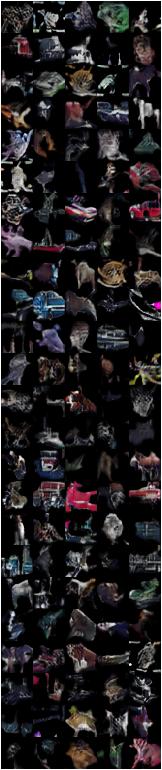}
\end{minipage}
\begin{minipage}{0.138\linewidth}
\center
\includegraphics[scale=0.33]{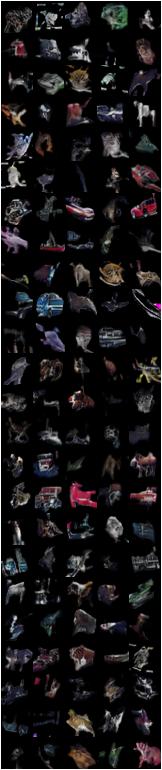}
\end{minipage}
\begin{minipage}{0.138\linewidth}
\center
\includegraphics[scale=0.33]{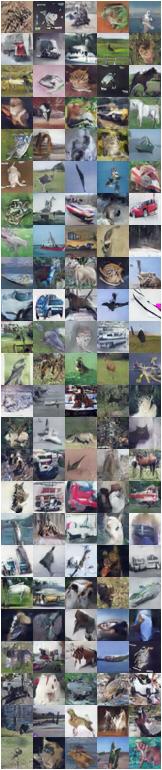}
\end{minipage}
\caption{Conditional generation results of our model on CIFAR-10. From left to right, the blocks are: real images, generated background images, foreground images, foreground masks, foreground images carved out by masks, carved foreground images after spatial transformation, and final composite (reconstructed) images.}
\label{Fig_CIFAROutputs_Conditional}
\end{figure}

Considering our model can generate object-like masks (shapes) for images, we conducted an experiment to evaluate whether our model can be potentially used for image segmentation and object detection. We make some changes to the model. For the background generator, the input is a real image instead of a random vector. Then the image is passed through an encoder to extract the hidden features, which replaces the random vector $\bm{z}_0$ and are fed to the background generator. For the foreground generator, we subtract the image generated by the background generator from the input image to obtain a residual image. Then this residual image is fed to the same encoder to get the hidden features, which are used as the input for foreground generator. In our conditional model, we want to reconstruct the image, so we add a reconstruction loss along with the adversarial loss. We train this conditional model on CIFAR-10. The (intermediate) outputs of the model is shown in Fig.~\ref{Fig_CIFAROutputs_Conditional}. Interestingly, the model successfully learned to decompose the input images into background and foreground. The background generator tends to do an image inpainting by generating a complete background without object, while the foreground generator works as a segmentation model to get object mask from the input image.

\begin{figure}[t]
\begin{minipage}{0.138\linewidth}
\center
\includegraphics[scale=0.33]{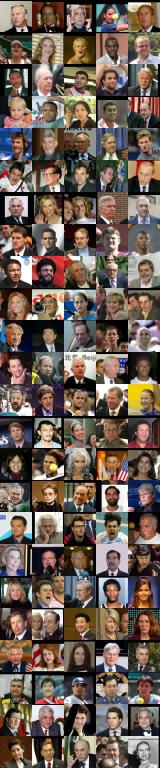}
\end{minipage}
\begin{minipage}{0.138\linewidth}
\center
\includegraphics[scale=0.33]{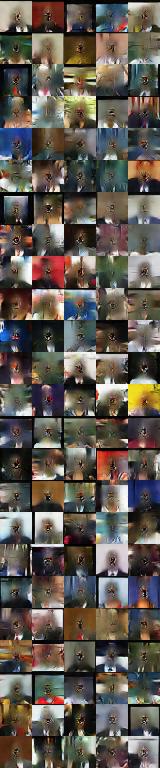}
\end{minipage}
\begin{minipage}{0.138\linewidth}
\center
\includegraphics[scale=0.33]{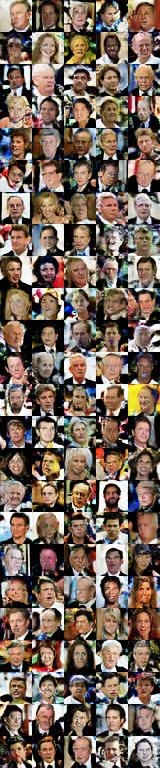}
\end{minipage}
\begin{minipage}{0.138\linewidth}
\center
\includegraphics[scale=0.33]{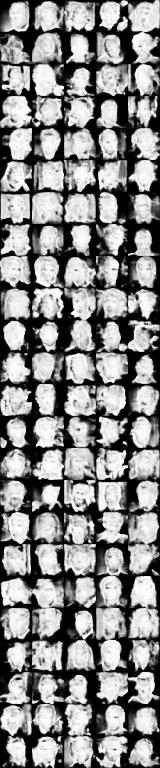}
\end{minipage}
\begin{minipage}{0.138\linewidth}
\center
\includegraphics[scale=0.33]{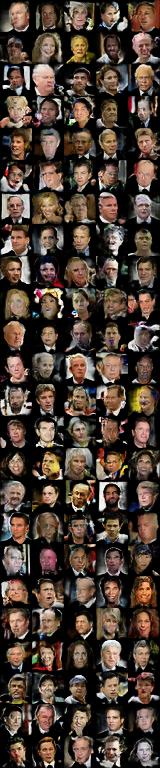}
\end{minipage}
\begin{minipage}{0.138\linewidth}
\center
\includegraphics[scale=0.33]{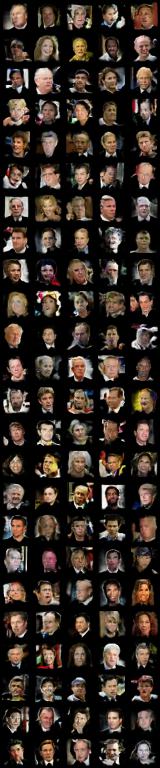}
\end{minipage}
\begin{minipage}{0.138\linewidth}
\center
\includegraphics[scale=0.33]{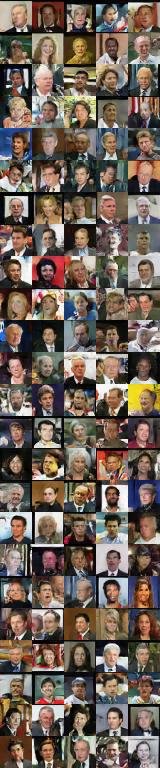}
\end{minipage}
\caption{Conditional generation results of our model on LFW, displayed with the same layout to Fig.~\ref{Fig_CIFAROutputs_Conditional}.}
\label{Fig_LFWOutputs_Conditional}
\end{figure}

Similarly, we also run the conditional LR-GAN on LFW dataset. As we can see in Fig.~\ref{Fig_LFWOutputs_Conditional}, the foreground generator automatically and consistently learned to generate the face regions, even though there are large portion of background in the input images. In other words, the conditional LR-GAN successfully learned to detection faces in images. We suspect this success is due to that it has low cost for the generator to generate similar images, and thus converge to the case that the first generator generate background, and the second generator generate face images.

Based on these experiments, we argue that our model can be possibly used for image segmentation and object detection in a generative and unsupervised manner. One future work would be verifying this by applying it to high-resolution and more complicate datasets.
\label{subsec_CondGeneration}

\end{document}